\documentclass{article}

\PassOptionsToPackage{numbers, sort, compress}{natbib}

\usepackage[final]{neurips_2025}

\usepackage[utf8]{inputenc} 
\usepackage[T1]{fontenc}    
\usepackage{url}            
\usepackage{booktabs}       
\usepackage{amsfonts}       
\usepackage{nicefrac}       
\usepackage{microtype}      

\usepackage[table,dvipsnames]{xcolor}
\usepackage{xspace}
\usepackage[colorlinks,citecolor=NavyBlue]{hyperref}       
\usepackage{inwoo}
\usepackage[accept]{editing}

\newcommand{\ourtitle}{From Black-box to Causal-box: \\Towards Building More Interpretable Models}

\title{\ourtitle}

\usepackage{titletoc}
\newcommand\DoToC{%
  \setcounter{tocdepth}{2} %
  \startcontents
  {\centering\Large\bfseries Appendix\par}
  \vskip3pt\hrule\vskip5pt
  \printcontents{}{1}{}
  \vskip3pt\hrule\vskip5pt
}

\author{%
Inwoo Hwang \qquad Yushu Pan \qquad Elias Bareinboim \\
Causal Artificial Intelligence Lab \\
Columbia University \\
\texttt{inwoo.hwang@columbia.edu \quad \{yushupan, eb\}@cs.columbia.edu}
}

\begin{document}

\maketitle

\begin{abstract}
Understanding the predictions made by deep learning models remains a central challenge, especially in high-stakes applications. A promising approach is to equip models with the ability to answer counterfactual questions -- hypothetical ``what if?'' scenarios that go beyond the observed data and provide insight into a model reasoning. In this work, we introduce the notion of causal interpretability, which formalizes when counterfactual queries can be evaluated from a \xst{model} \xadd{specific class of models} and observational data. We analyze two common model classes -- blackbox and concept-based predictors -- and show that neither is causally interpretable in general. To address this gap, we develop a framework for building models that are causally interpretable by design. Specifically, we derive a complete graphical criterion that determines whether a given model architecture supports a given counterfactual query. This leads to a fundamental tradeoff between \xadd{causal} interpretability and predictive accuracy, which we characterize by identifying the unique maximal set of features that yields an interpretable model with maximal predictive expressiveness. Experiments corroborate the theoretical findings. 

\end{abstract}

\section{Introduction}
\label{sec:intro}
Despite the remarkable success of deep learning models across a wide range of tasks -- including image recognition \citep{krizhevsky2012imagenet, he2016deep}, natural language processing \citep{vaswani2017attention, devlin2018bert}, and reinforcement learning \citep{sutton1998reinforcement,silver2016mastering} -- these models remain fundamentally opaque. 
Although they are highly effective at predicting labels based on statistical correlations in the data, they lack the capacity to explain the reasoning behind their predictions, earning them the colloquial label of ``black boxes.''
In other words, current models are difficult to interpret: they lack the ability to justify why a particular decision was made, identify which input factors were most influential, or reason about how outcomes might differ under alternative, counterfactual conditions.
This interpretability gap raises concerns in high-stakes domains such as healthcare, law, and scientific discovery, where understanding how and why a model makes a decision is as important as the decision itself. 

A rich body of research on explainable AI (XAI) has been developed to better understand the behavior of learned models. For instance, post-hoc explanation methods such as LIME~\citep{ribeiro2016should}, SHAP~\citep{lundberg2017unified}, and Grad-CAM~\citep{selvaraju2017grad} generate local or visual attributions in terms of pixels or extracted features to help interpret predictions.
Other approaches aim to build intrinsically interpretable models, such as those that impose sparsity constraints~\citep{ng2011sparse}, restrict final layers~\citep{wong2021leveraging}, or leverage decision tree structures~\citep{wan2020nbdt}, often trading off model complexity for greater transparency.
While these techniques offer useful insights, they fail to bridge the gap between low-level features and high-level, human-understandable features that might explain the behavior of a model.

One promising avenue for bridging this gap is counterfactual reasoning. Answering what if questions -- such as ``Would the diagnosis have changed if a different treatment had been administered?'' or ``Would the person have been classified differently if their income were higher?'' -- plays a central role in human reasoning and forms the basis of many explanatory and decision-making processes \citep{pearl2009causality, pearl2018book, bareinboim2022pearl}.
Enabling AI systems to reason counterfactually opens the door to more interpretable models -- ones that can not only predict outcomes accurately but also explain their decisions in a meaningful, human-aligned way.

Recently, concept-based prediction models \citep{koh2020concept,oikarinenlabel} have been proposed to improve interpretability by enabling reasoning over human-understandable features. These models aim to answer counterfactual queries of the form: ``Given an input $\mathbf{x}$, how would the model’s prediction change if a feature $\mathbf{W}$ were modified from $\mathbf{w}$ to $\mathbf{w'}$?''
Such queries allow users to explore the influence of high-level features -- like the presence of a smile or the existence of a tumor -- on a model’s prediction, providing a possible route to assess whether the model reasoning aligns with human expectations.

Despite their appeal, existing concept-based approaches are oblivious to the causal relationships between features. 
As a result, they may not reflect the real-world mechanisms or incorporate common-sense knowledge faithfully.
While some recent methods attempt to introduce causal structure into concept-based models~\citep{dominici2024causal}, they frequently lack guarantees of counterfactual consistency -- that is, the property that models within the exact class yield consistent answers to the same counterfactual query.

\begin{figure}
\centering
\includegraphics[width=0.9\linewidth]{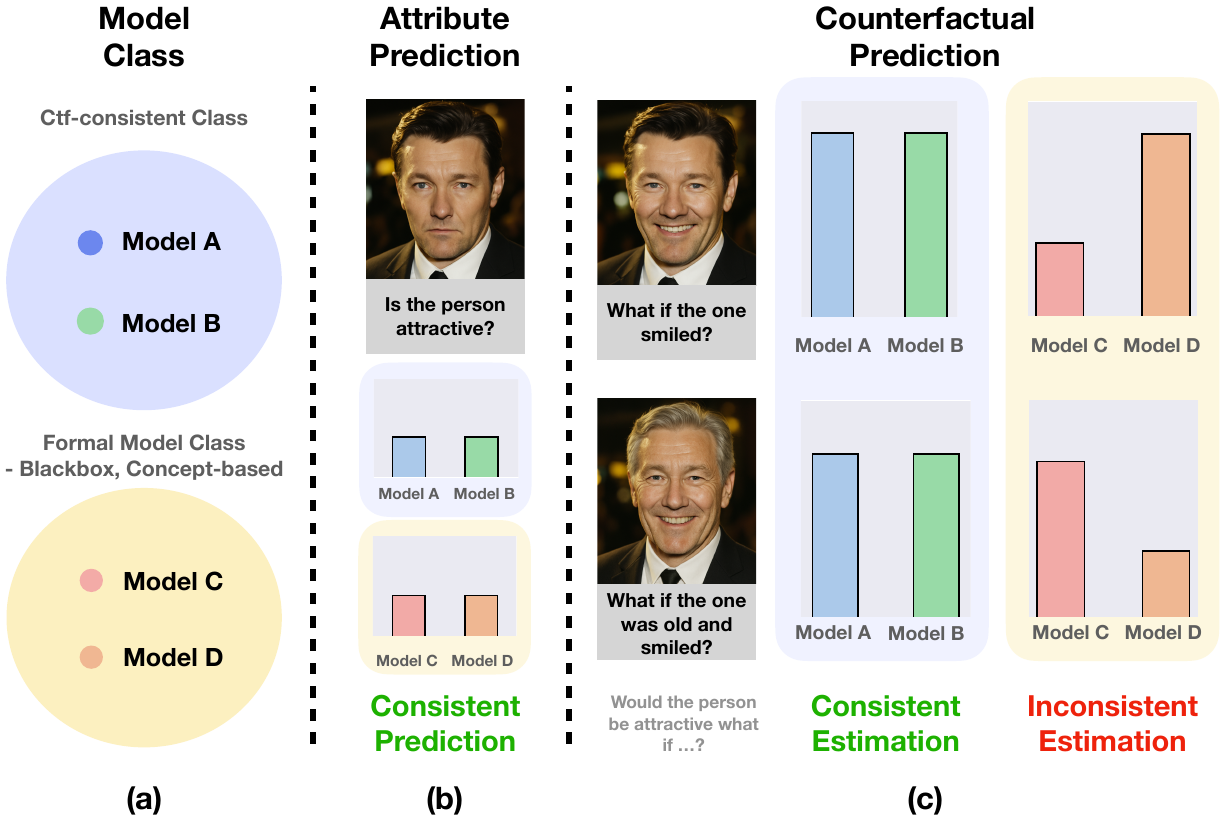}
\caption{
(a) Illustration of different model classes: counterfactually consistent models (blue) and blackbox/concept-based models (yellow).
(b) Original input image and corresponding predictions from each model.
(c) Counterfactual predictions: models in the top row predict consistently across instantiations within the class, while those in the bottom row produce inconsistent predictions.
}
\label{fig:teaser}
\end{figure}

To illustrate this limitation, consider a task of predicting facial attractiveness.
Suppose two models, C and D, from the same concept-based class, represented by the yellow circle in \Cref{fig:teaser}-(a), are trained on the same dataset. 
They first will have the identical attribute prediction, for example, both will predict a lower attractiveness score for the given image (\Cref{fig:teaser}-(c), yellow). 
However, when they evaluate the counterfactual question ``What would the attractiveness be had the person smiled?'', model C will maintain the low attractiveness score while model D will raise the attractiveness score (\Cref{fig:teaser}-(c), yellow). 
This discrepancy reveals a deeper issue: the model class is not counterfactually interpretable, as it does not constrain the space of counterfactual responses.
In such cases, users have no principled way to determine which answer to trust, rendering the query effectively unanswerable. 
In contrast, the model class in blue is desirable since any pair of models -- such as model A and B -- will give the exact same answer for both attribute and counterfactual predictions. 
In this case, one can assert that the attractiveness would be raised had the person smiled, which indicates the model made the decision based on the feature ``Smile'' and this is aligned with human understanding \citep{horn2021smile}.

In this work, we introduce the notion of causal interpretability, which concerns whether a prediction model can be interpreted consistently across counterfactual scenarios -- drawing a connection between XAI and causal inference \citep{pearl2009causality, bareinboim2022pearl}.
Intuitively, a model class is said to be \emph{causally interpretable} if all models within the class yield consistent predictions under counterfactual interventions, as illustrated in blue in \Cref{fig:teaser}.
We then show that a blackbox model, which maps inputs directly to labels, is never causally interpretable.
That is, such models fundamentally lack the structure needed to answer counterfactual questions.
We also demonstrate theoretically that concept-based models \citep{koh2020concept}, which rely on all observed features for prediction, are also not guaranteed to be causally interpretable.
\xadd{Interestingly, we show that causal interpretability can be recovered by constraining the model to use only a certain subset of features.}

Against this background, we develop a general approach for building causally interpretable models that can answer counterfactual queries \xadd{consistently} by design.
Specifically, we propose a complete graphical criterion for determining whether a model that uses a given set of features for prediction is causally interpretable with respect to a counterfactual query.
This enables the understanding of (i) which counterfactual questions a given model can answer, and (ii) which models can answer a given counterfactual question.
Our framework also reveals a fundamental tradeoff between causal interpretability and predictive accuracy. We characterize the unique maximal set of features that preserves causal interpretability, thereby providing a principled method for building models with maximal expressive power under interpretability constraints.
A notable practical implication is that our approach does not require full specification of the causal graph or modeling of unobserved confounders; it only involves the descendants of the target features in the counterfactual query.
Experimental results corroborate the proposed theory.
More specifically, our contributions are as follows:
\begin{itemize}[leftmargin=*]
\item (\Cref{sec:causal-interpretability}) We introduce the notion of causal interpretability (\Cref{def:causal-interpretability}), which states whether we can evaluate the prediction of the model under counterfactual conditions from observational data.
Based on this formulation, we show that a blackbox model is never interpretable (\Cref{prop:bp-noninterpretability}), whereas a concept-based model \xst{is also often} \xadd{is also} not interpretable \xadd{in general}, in contrast to prior belief.
\item (\Cref{sec:generalized}) We develop a graphical criterion that determines whether the model is \xadd{causally} interpretable with respect to the query (\Cref{thm:graphicalcriteria}). We characterize the unique maximal set of features yielding interpretable architecture (\Cref{prop:uniqueness}) and provide a practical way of evaluating such queries from the data (\Cref{thm:closed-form}). Finally, these results reveal a fundamental tradeoff between the causal interpretability and predictive accuracy (\Cref{thm:tradeoff}).
\end{itemize}

\label{sec:preliminary}
\paragraph{Preliminary.} 
Here, we introduce notations and terminologies used in the paper.
We use bold letters to denote a set of random variables or their assignments. We use capital letters to denote a random variable or a random vector (e.g., $\rmX$) and lower case letters to denote their assignments (e.g., $\rvx$). 
\xadd{$\rvx \cup \rmZ$ denotes the subset of $\rvx$ corresponding to variables in $\rmZ$ and $\rvx \setminus \rmZ$ denotes the value of $\rmX \setminus \rmZ$ consistent with $\rvx$.}

We employ a structural causal model \citep{pearl2009causality,bareinboim2022pearl} as our semantical framework.
A structural causal model (SCM) $\gM$ is a 4-tuple $\langle \rmU, \rmV, \gF, P(\rmU)\rangle$, where $\rmU$ is a set of exogenous variables, $\rmV=\{V_1, \cdots, V_n\}$ is a set of endogenous variables, $\gF=\{f_{V_1}, \cdots f_{V_n}\}$ is a set of functions determining $\rmV$ as $V_j \gets f_{V_j}(\rmPa_{V_j}, \rmU_{V_j})$, where $\rmPa_{V_j} \subseteq \rmV\setminus \{V_j\}$ and $\rmU_{V_j} \subseteq \rmU$ for all $V_j\in \rmV$, and $P(\rmU)$ is a distribution over $\rmU$.
An SCM $\gM$ induces a causal diagram $\gG$ and a distribution over the endogenous $P(\rmV)$.
We use graphical kinship to represent the relationships between the variables. 
\xadd{$ND(W)$ denotes non-descendants of a variable $W$, and $ND(\rmW) \coloneq \cap_{W_i \in \rmW} ND(W_i)$ denotes non-descendants of a set of variables $\rmW$.}
We now define an SCM that describes a generative process that includes images $\rmX$ and labels prediction $\hY$ \citep{pan2024counterfactual}.
\begin{definition}[Augmented SCM]
\label{def:ascm}
An augmented SCM (ASCM) over a generative level SCM $\gM_{0} = \langle \*U_0, \*V_0, \gF_0, P^0(\*U_{0}) \rangle$ is a tuple $\gM = \langle \*U, \{\*V, \*X, \hY\}, \gF, P(\*U) \rangle$ such that \\
(1) exogenous variables $\*U = \{\*U_{0}, \*U_{\*X}\}$; \\ 
(2) $\*V  = \*V_0$ are labeled observed endogenous variables, $\*X$ is an $m$-dimensional mixture variable, and $\hY$ is a (predicted) label; \\
(3) $\gF = \{\gF_{0}, f_{\*X}, f_\hY\}$,  
where $f_{\*X}$ maps from (the respective domains of) $\*V \cup {\*U}_{\*X}$ to $\*X$
and a classifier $f_{\hY}$ maps from (the respective domains of) the subset of $\{\rmV, \rmX\}$ to $\hY$; and
\\
(4) $P(\*U_0) = P^0(\*U_0)$.
\end{definition}
An ASCM $\gM$ represents a sequential generative procedure of latent generative factors (i.e., concepts) $\rmV$, the image $ \rmX$, and the label prediction $\hY$.
First, the latent features $\rmV$ are generated by the underlying $\gM_0$.
The induced causal diagram $\gG_\rmV$ is called a latent causal graph (LCG). %
The high-dimensional mixture $\rmX$ (e.g., image) is then generated from $\rmV$ (and $\rmU_\rmX$), and subsequently, $\hY$ is generated from the subset of $\{\rmV, \rmX\}$, where $f_\hY$ is a classifier that predicts the label.
We let $\Omega \coloneqq \{\gM: \text{ASCM over } \gM_0\}$ be the space of ASCMs. 
Omitted proofs are provided in \Cref{appendix:theory-proofs}.

\section{Causal Intepretability -- Foundations} \label{sec:causal-interpretability}

In this section, we formalize the notion of causal interpretability and examine whether existing approaches could elicit counterfactual questions \xadd{consistently} in a valid manner.

\begin{figure}[t!]
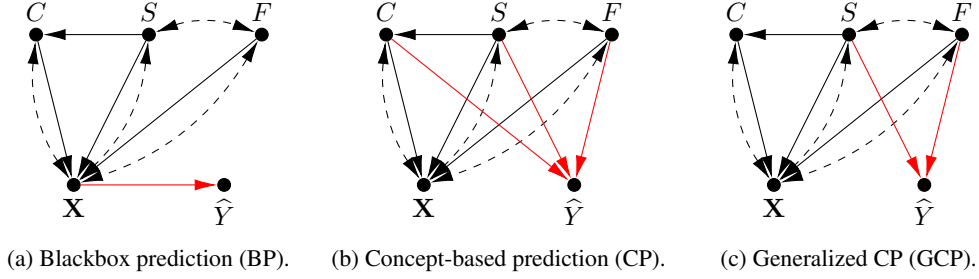

\centering
\begin{subfigure}{.32\linewidth}\centering
\begin{tikzpicture}[SCM]
\input{fig/base-scm-node3}
\path [red] (X) edge (Y);
\end{tikzpicture}
\caption{Blackbox prediction (BP).}
\label{fig:comparison-bp}
\end{subfigure}
\hfil
\begin{subfigure}{.32\linewidth}\centering
\begin{tikzpicture}[SCM]
\input{fig/base-scm-node3}
\path [red] (1) edge (Y);
\path [red] (2) edge (Y);
\path [red] (3) edge (Y);
\end{tikzpicture}
\caption{Concept-based prediction (CP).}
\label{fig:comparison-cp}
\end{subfigure}
\hfil
\begin{subfigure}{.32\linewidth}\centering
\begin{tikzpicture}[SCM]
\input{fig/base-scm-node3}
\path [red] (2) edge (Y);
\path [red] (3) edge (Y);
\end{tikzpicture}
\caption{Generalized CP (GCP).}
\label{fig:comparison-gcp}
\end{subfigure}
\caption{Causal diagrams for different types of predictive models.}
\label{fig:comparison}
\vspace{-7pt}
\end{figure}

We start by analyzing two important classes of predictive models: \textit{blackbox} and \textit{concept-based} models.
As illustrated in \Cref{fig:comparison-bp}, blackbox prediction (BP) models make a prediction on the label from the image \xadd{pixels} $\rmX$ (i.e., $f_\hY: \gD(\rmX) \to \gD(\hY)$). 
\xst{This means that a blackbox does not have access to any of the causal factors that generated the data.}
In contrast, concept-based prediction (CP) models predict the label based on the generative factors of the image (i.e., $f_\hY: \gD(\rmV) \to \gD(\hY)$), as illustrated in \Cref{fig:comparison-cp}.
In other words, the classifier of a concept-based model uses the features to make the predictions, instead of the image itself.
Formally, a class of BP models and a class of CP models are respectively denoted as $\Omega_\text{BP}$ and $\Omega_\text{CP}$, where
$\Omega_\text{BP} \coloneqq \{\gM \in \Omega \mid f_\hY: \gD(\rmX) \to \gD(\hY) \}$ and 
$\Omega_\text{CP} \coloneqq \{\gM \in \Omega \mid f_\hY: \gD(\rmV) \to \gD(\hY) \}$.
The following examples illustrate the generative process of BP and CP models.

\begin{example}[Blackbox Model]
\label{ex:bp}

Consider a task of estimating the attractiveness of a human face represented in an image $\*X$. 
Augmented generative process (ASCM) of the prediction by a BP model is given as $\cM_{\text{BP}} = \langle \*U=\{U_F, U_{S}, U_{C_1}, U_{C_2}, \*U_{\*X}\}, \{\{F, S, C\}, \*X, \widehat{Y}\}, 
 \cF^{\text{BP}}, P^{\text{BP}}(\*U) \rangle$, where
\begin{equation}
\label{eq:bp}
{\cF^{\text{BP}}} = \left\{
\begin{aligned}
F & \leftarrow U_F \oplus U_S \\
S & \leftarrow U_S \\
C & \leftarrow (\neg S \land U_{C_1}) \oplus (S \land U_{C_2})  \\
\*X & \leftarrow {f}_{\*X}(F, S, C, \*U_{\*X}) \\
\widehat{Y} & \leftarrow f_{\widehat{Y}}(\*X),
\end{aligned}
\right.
\end{equation}
$\widehat{Y}$ is the label (attractiveness) prediction, the exogenous variables $U_F, U_S, U_{C_1}, U_{C_2}$ are independent binary variables, and $P^{\text{BP}}(U_F = 1) = 0.4, P^{\text{BP}}(U_{S} = 1) = 0.6, P^{\text{BP}}(U_{C_1} = 1) = 0.3, P^{\text{BP}}(U_{C_2} = 1) = 0.6$. The exogenous variable $\*{U}_{\mathbf{X}}$ (representing other generative factors) can include (or be correlated to) $\{U_F, U_S, U_{C_1}, U_{C_2}\}$. 
The causal diagram induced by $\cM_\text{BP}$ is shown in \Cref{fig:comparison-bp}.

In terms of prediction, the process of obtaining $\widehat{Y}$ has three steps. 
First, latent generative features $F$ (gender), $S$ (smiling), and $C$ (high cheekbones) are generated.
Then, $f_{\mathbf{X}}$ maps the observed generative features $\{F, S, C\}$ and unobserved generative factors $\*U_{\*X}$ to the images $\*X$ in the pixel levels.
Finally, the predictor $f_{\widehat{Y}}$ takes these pixels as input to estimate $\widehat{Y}$ in the corresponding model. 
The functions $f_{\mathbf{X}}$ and $f_{\widehat{Y}}$ can be aggregated as
$\widehat{Y} \leftarrow f_{\widehat{Y}} \circ f_{\mathbf{X}}(F, S, C, \mathbf{U}_{\mathbf{X}})$.
This illustrates that the prediction of $\widehat{Y}$ by a BP model is made based on all observed features $\{F, S, C\}$ and unobserved features $\mathbf{U}_{\mathbf{X}}$.
\hfill $\blacksquare$
\end{example}
\begin{example}[Concept-based Model]
\label{ex:cp}
The main difference between the class of CP models $\Omega_{\text{CP}}$ and the class of BP models $\Omega_{\text{BP}}$ is the form of the classifier $f_{\widehat{Y}}$. 
Consider the same generative process of observed features $\*V_0 = \{F, S, C\}$\footnote{In practice, the annotations of the features are provided in many real-world datasets across various domains, e.g., human face \citep{liu2018large}, medical images \citep{nevitt2006osteoarthritis}, and animal species \citep{wah2011caltech}. Otherwise, the common practice is to extract their annotations with vision-language models \citep{radford2021learning}, which is shown to be effective \citep{oikarinenlabel,yang2023language}.} and the image $\*X$ in \Cref{ex:bp}. 
Let us consider a CP model $\cM_{\text{CP}} = \langle \*U=\{U_F, U_{S}, U_{C_1}, U_{C_2}, \*U_{\*X}\}, \{\{F, S, C\}, \*X, \widehat{Y}\}, 
\cF^{\text{CP}}, P^\text{CP}(\*U) \rangle$, where
the generative process of $F, S, C, \rmX$ is the same as \Cref{eq:bp}, $\hY$ is generated as 
\begin{equation}
\label{eq:cp}  
{\widehat{Y}}  \leftarrow {f_{\widehat{Y}}(F, S, C)},
\end{equation}
and $P^\text{CP}(\*U)$ is equal to $P^\text{BP}(\*U)$ in \Cref{ex:bp}.
In words, this means that instead of predicting $\widehat{Y}$ based on pixels (i.e., image $\*X$), the classifier $f_{\widehat{Y}}$ directly predicts $\widehat{Y}$  based on observed features  
$F, S, C$. 
The causal diagram induced by $\cM_\text{CP}$ is shown in \Cref{fig:comparison-cp}.
\hfill $\blacksquare$
\end{example}

\Cref{ex:bp,ex:cp} illustrate two different types of predictive models, where the classifier predicts the label directly from the image $\rmX$ (i.e., $\Omega_\text{BP}$) or from the generative features $\rmV$ (i.e., $\Omega_\text{CP}$). 
While both types have showcased their capability to achieve reasonably high predictive accuracy in many domains \citep{krizhevsky2012imagenet,koh2020concept,oikarinenlabel,he2016deep,yuksekgonul2023posthoc,kim2023probabilistic,jeon2025locality,dominici2025counterfactual,sun2025concept,ismail2025concept}, 
it is unclear at this moment whether 
we can interpret how they would predict under counterfactual scenarios, such as 
``\textit{how attractive the person would be had the one been smiling?}''.
The following notion of \textit{causal interpretability} formally states whether the counterfactual questions can be answered from the model.

\begin{figure}[t]
\centering
\includegraphics[width=0.8\linewidth]{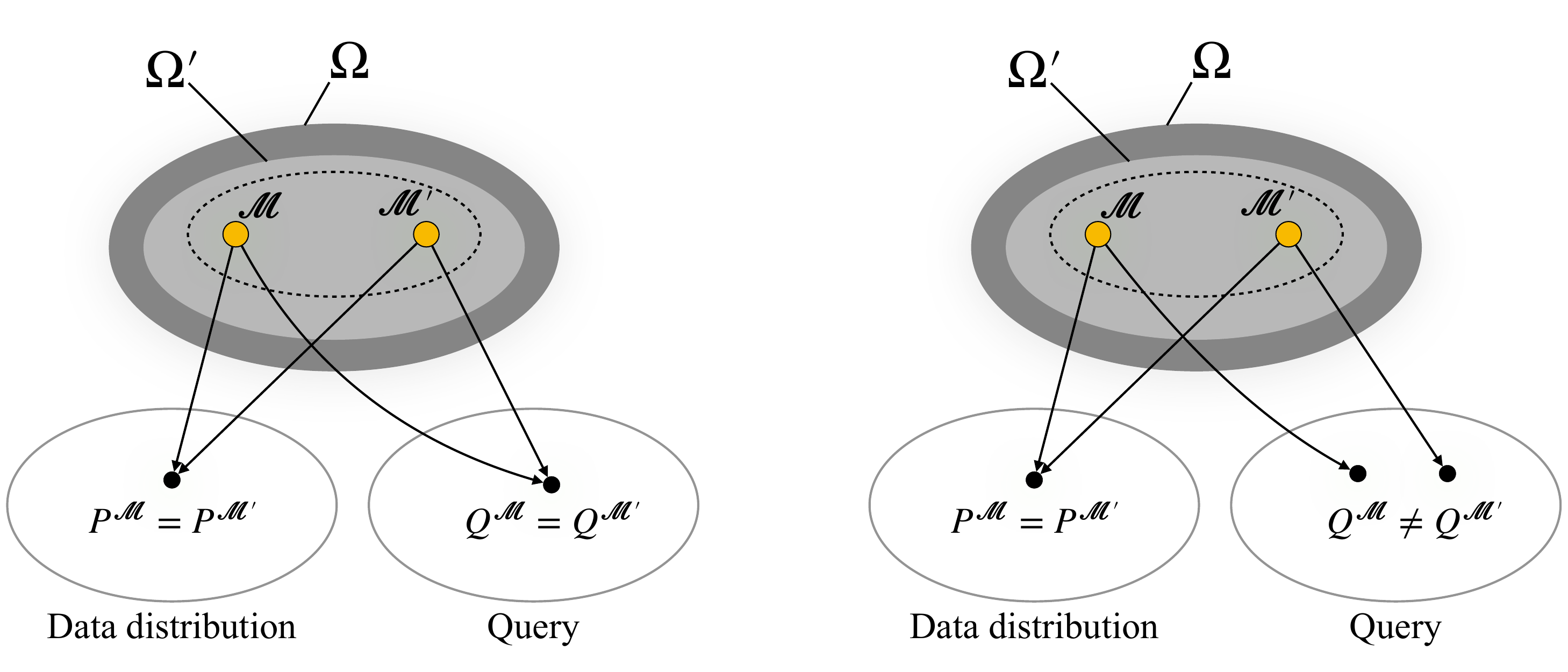}
\caption{(\textbf{Left}) $\Omega'$ is causally interpretable if a query can be uniquely computed from the observational data.  (\textbf{Right}) A query cannot be uniquely computed from the observational data if $\Omega'$ is not causally interpretable.}
\label{fig:causal-interpretability}
\end{figure}

\begin{definition}[Causal Interpretability]
\label{def:causal-interpretability}
Consider a specific model class $\Omega'\subset\Omega$, where $\Omega$ is the space of ASCMs. We say the class $\Omega'$ is \textbf{causally interpretable w.r.t. a query $Q$} if $Q^{\gM_1}=Q^{\gM_2}$ for $\forall \gM_1, \gM_2 \in \Omega'$ s.t. $P^{\gM_1}(\rmV, \rmX, \hY) = P^{\gM_2}(\rmV, \rmX, \hY)$.  %
\end{definition}
In words, $\Omega'$ denotes a certain design choice of the models for predicting the label,
that is, it is a space of prediction model candidates \xadd{(i.e., model class)}.
$\Omega'$, for instance, can be $\Omega_\text{BP}$, when we want to predict the label directly from the image (\Cref{fig:comparison-bp}), or $\Omega_\text{CP}$, when the classifier uses all observed features (\Cref{fig:comparison-cp}). 
For a query $Q$, we are concerned with the counterfactual questions such as ``\textit{What if the person had smiled?}'', which is written in counterfactual notion as $P(\hY_{S=1} \mid \rmX=\rvx)$, and more generally as 
$Q(\rmW) \coloneqq P(\widehat{Y}_\rmW \mid \rmX)$.\footnote{Note that the definition is general in terms of the query $Q$, which could vary across different domains, e.g., natural direct effect in fairness analysis \citep{plecko2022causal}.}

In other words, the notion of causal interpretability states whether one can understand the behavior of the model under different counterfactual conditions.
If the model is causally interpretable, the counterfactuals can be evaluated from the observational data (\Cref{fig:causal-interpretability}, left). Otherwise, the model fundamentally cannot answer the counterfactual question from observational data, and thus, we cannot interpret their behavior under counterfactual scenarios (\Cref{fig:causal-interpretability}, right).
We now analyze two types of predictive models discussed above (i.e., BP model in \Cref{ex:bp} and CP model in \Cref{ex:cp}) and examine their causal interpretability, i.e., whether they can evaluate counterfactuals from observational data. 

\begin{example}[Continued from \Cref{ex:bp}]
\label{ex:bp-noninterpretability}
Consider the BP model $\cM_\text{BP}$ in \Cref{ex:bp}. {Let $\*U_{\*X}$ includes} another independent variable \xadd{$U_S$}, namely, $\*U_{\*X} = \{U_S, \mathbf{U}^{-}_{\mathbf{x}}\}$, \xadd{where $U_S\perp \*U \setminus U_S$}; let the observational quantity 
$P(F=0, S=1, C=1 \mid \mathbf{X}={\mathbf{x}}) = 1$,
which means that the face is 
of 
a male ($F=0$), who is smiling ($S=1$), and with the cheekbones high ($C=1$), given in an image $\rmX=\rvx$.
The generative process of $\hY$ is as 
$\widehat{Y} \leftarrow f_{\widehat{Y}} \circ f_{\mathbf{X}}(F, S, C, \*U_\*X) = \mathbf{1}[S > 0.5]$.

Consider another BP model $\cM_{\text{BP}}'$ with the same generative process of $\cM_\text{BP}$, but for in $\cM_{\text{BP}}'$, the classifier $f^{'}_{\widehat{Y}}$ is given by:
$\widehat{Y} \leftarrow f_{\widehat{Y}} \circ f_{\mathbf{X}} (F, S, C, \*U_\*X) = \mathbf{1}[U_S > 0.5]$.
Since $S = U_S$,  the two BP models 
$\cM_{\text{BP}}$ and $\cM_{\text{BP}}'$ agrees with the observational data, i.e., $P^{\cM_\text{BP}}(\mathbf{V}, \mathbf{X}, \widehat{Y}) = P^{\cM_{\text{BP}}'}(\mathbf{V}, \mathbf{X}, \widehat{Y})$,
which will lead to the same predictions (and corresponding accuracy).

Now, consider the counterfactual quantity
"Given the image $\rmX=\rvx$, would the prediction still be attractive ($\hY=1$) had the person not smiled ($S=0$)?", namely, $Q(S) = P(\widehat{Y}_{S=0} = 1 \mid \mathbf{X} = \mathbf{x})$. Intuitively, a smaller value of $P(\widehat{Y}_{S=0} = 1 \mid \mathbf{X} = \mathbf{x})$ implies the model is more reliable since changing a face to non-smiling reduces the attractiveness in general based on common sense knowledge \citep{horn2021smile}. 
For the first BP model ${\cM_{\text{BP}}}$, $Q(S)$ evaluates as 
$P^{\cM_{\text{BP}}}(\widehat{Y}_{S=0} = 1 \mid \mathbf{X}= \mathbf{x}) = \mathbf{1}[S = 0 > 0.5] = 0$.
However, for the second BP model $\gM'_\text{BP}$, $Q(S)$ evaluates as 
$P^{\cM_{\text{BP}}'}(\widehat{Y}_{S=0} = 1 \mid \mathbf{X}= \mathbf{x}) = \mathbf{1}[U_S = 1 > 0.5] = 1$.
Details for these derivations are provided in \Cref{appendix:theory}.

Note that each BP model evaluates the counterfactual query in a completely different way, and the two models are somewhat inconsistent.
In practice, if one chooses the class of BP models $\Omega_\text{BP}$ for this prediction task, 
the above counterfactual question cannot be answered correctly, since two BP models can give an exact opposite answer even if the two models agree perfectly with the observational distribution and their predictions. 
\xadd{In other words, the blackbox model class cannot answer counterfactual $Q(S)$ consistently from observational data, and its behavior cannot be interpreted under corresponding counterfactual conditions.}
\hfill $\blacksquare$
\end{example}

One may surmise that \Cref{ex:bp-noninterpretability} is a pathological case, which for some reason does not allow the evaluation of counterfactual queries in a consistent manner. The next result shows that this is not the case for an arbitrary query $Q(\rmW)$ and a latent causal graph $\gG_\rmV$.

\begin{restatable}[Non-interpretability of BP]{proposition}{bpnoninterpretability}
\label{prop:bp-noninterpretability}
For any latent causal graph $\gG_\rmV$,
$\Omega_\text{BP}$ is not \xadd{causally} interpretable w.r.t. $Q(\rmW)$ for any $\rmW\subseteq \rmV$.
\end{restatable}

Given this impossibility results for the class of blackbox models, one may be tempted to believe that 
a CP architecture is causally interpretable, as it predicts the label directly from the features where the unobserved factors $\*U_{\*X}$ are filtered out. However, the following illustrates that this is not the case.

\begin{example}[Continued from \Cref{ex:cp}]
\label{ex:cp-noninterpretability}
Consider the CP model $\cM_\text{CP}$ in \Cref{ex:cp}. Similar to \Cref{ex:bp-noninterpretability}, consider an  observational quantity $P(F=0, S=1, C=1 \mid \mathbf{X}={\mathbf{x}}) = 1$.
$\hY$ is generated as follows:
\begin{align}
    \label{eq:agg-cp-nonid-1}
    \widehat{Y} \leftarrow f_{\widehat{Y}} (F, S, C) = \mathbf{1}[S + C > 0.5].
\end{align}
Now consider another CP model $\cM'_\text{CP}$ that is the same as $\cM_\text{CP}$, except for
$C \leftarrow f'_{C} (S, U_{C_1}) = (S \lor U_{C_1})\land U_{C_2}$
and $P(U_{C_1} = 1) = 0.5$. 
We have $P^{\cM_\text{CP}}(\*V, \*X, \widehat{Y}) = P^{\cM_{\text{CP}}'}(\*V, \*X, \widehat{Y})$ and $\cM_{\text{CP}}'$ is compatible with the graphical constraints in \Cref{fig:comparison-cp}. 
Now consider the same counterfactual quantity $P(\widehat{Y}_{S=0} = 1 \mid \mathbf{X} = \mathbf{x})$ in \Cref{ex:bp-noninterpretability}. 
For $\cM_{\text{CP}}$, we have 
$P^{\cM_{\text{CP}}}(\widehat{Y}_{S=0} = 1 \mid \mathbf{X}= \mathbf{x}) = P^{\cM_{\text{CP}}}(C_{S=0}=1 \mid F=0, S=1, C=1) = 0.3$.
However, for the second CP model, 
$P^{\cM_{\text{CP}}'}(\widehat{Y}_{S=0} = 1 \mid \mathbf{X}= \mathbf{x}) = P^{\cM_{\text{CP}}'}(C_{S=0}=1 \mid F=0, S=1, C=1)= 0.5$.
This implies that the two CP models are also inconsistent w.r.t $Q(S)$. In other words, even prediction using features $\*V$, not pixels $\*X$, counterfactual queries induced by the CP models can still differ from each other. \hfill $\blacksquare$
\end{example}

\section{A Causal Approach Towards More Interpretable Models}
\label{sec:generalized}
In this section, we establish a principled way of understanding causal interpretability from a graphical point of view and propose a generalized framework for building causally interpretable models.

\subsection{Generalized Concept-based Models}
We first define generalized concept-based prediction (GCP) models, a broader class that predicts the label from an arbitrary set of observed features.

\begin{definition}[Generalized Concept-based Prediction]
\label{def:generalized-concept-model}
Let $\rmT\subseteq \rmV$ be a set of features that is used as a predictor of the label.
That is, a classifier $f_\hY$ makes a prediction on a label based on $\rmT$. We say such predictive models as generalized concept-based models. A class of GCP models that employ the features $\rmT$ for prediction is denoted as
$\Omega_{\textnormal{GCP}(\rmT)} \coloneqq \{\gM \in \Omega \mid f_\hY: \gD(\rmT) \to \gD(\hY) \}$.
\end{definition}

Compared to CP models, GCP models employ a selected set of features $\*T \subseteq \*V$ as a predictor of the label, which relaxes the requirement of CP where all features are considered.

The selection of the features $\rmT$ in a GCP model should be specified during the model building stage,
and our goal is to understand the implications of different choices of $\rmT$ and which ones could lead to causally interpretable models (i.e., satisfying \Cref{def:causal-interpretability}).
To answer this question systematically, we introduce a graphical criterion for determining whether a model satisfies causal interpretability.
\begin{restatable}[Graphical Criterion]{theorem}{graphicalcriteria}
\label{thm:graphicalcriteria}
Consider GCP models that employ a set of features $\rmT$ as a predictor of the label.
$\Omega_{\textnormal{GCP}(\rmT)}$ is causally interpretable w.r.t. a query $Q(\rmW)$ if and only if $\rmT\subseteq \rmW\cup ND(\rmW)$. %
\end{restatable}

In words, this result says that a query $Q(\rmW)$ can be evaluated if the model uses the features among $\rmW$ or non-descendants of $\rmW$ to make a prediction on the label. In other words, the models that use any descendant of $\rmW$ cannot answer counterfactual question and no guarantee can be provided on how they would make predictions under the corresponding counterfactual scenarios.\footnote{Note that for the case of $\rmX = \rmT$, $\Omega_{\textnormal{BP}}$ is not interpretable w.r.t. any $Q(\rmW)$ since $\rmX$ is a descendant of $\rmW$ for any $\rmW\subseteq \rmV$, generalizing \Cref{prop:bp-noninterpretability}. Similarly, $\Omega_{\textnormal{GCP}(\rmT)}$ is also never interpretable if $\rmX \in \rmT$, i.e., \textit{hybrid} models that make predictions based on the combination of the image and features.}

\Cref{thm:graphicalcriteria} enables one to identify the architectures (associated with \(\*T\)) that are causally interpretable with respect to given counterfactual queries. Interestingly, the models 
that are potentially causally interpretable are not unique. The following formalizes the notion of admissible architectures.

\begin{definition}[T-Admissible Set]
\label{def:t-ad}
We say $\rmT$ is T-admissible w.r.t. $\rmW_\star=\{\rmW_1, \rmW_2, \cdots\}$ if $\Omega_{\textnormal{GCP}(\rmT)}$ is interpretable w.r.t. $Q(\rmW_i)$ for all $\rmW_i\in \rmW_\star$. A set of T-admissible sets w.r.t. $\rmW_\star$ is denoted as $\text{T-Ad}(\rmW_\star)$.
\end{definition}

To illustrate, T-admissible set represents model architectures that can answer (potentially multiple) counterfactual queries $Q(\rmW_1), Q(\rmW_2), \cdots$.
For example, in \Cref{fig:comparison}, 
eligible models that one can evaluate $Q(S)$
is GCP models whose classifier employs $\{S\}$, $\{F\}$, or $\{S, F\}$ as a predictor of the label, i.e.,
T-admissible set corresponds to the query $Q(\{S\})$ is $\text{T-Ad}(\{S\}) = \{\{S\}, \{F\}, \{S, F\}\}$. 

Given the multiplicity of admissible models, 
our goal is to find the models that use as many features as possible to predict the label $\hY$, i.e., maximal $\rmT$, as it would be beneficial in terms of predictive accuracy.
We denote it as a \textit{maximal T-admissible set}, which is formally defined below.

\begin{definition}[Maximal T-Admissible Set]
Suppose $\rmS\in \text{T-Ad}(\rmW_\star)$ and $\rmS' \not\in \text{T-Ad}(\rmW_\star)$ for any $\rmS' \supsetneq \rmS$.
We denote such $\rmS$ as $\text{Max-T-Ad}(\rmW_\star)$.
\end{definition}

In other words, a maximal T-admissible set is a T-admissible set that would cease to be T-admissible if any additional variable were added to it.
\xadd{Note that once a set is not T-admissible, adding more variables never makes it T-admissible again by \Cref{thm:graphicalcriteria}.}
Identifying a maximal T-admissible set would lead to a model with maximal predictive power while retaining causal interpretability.
One might suspect that multiple maximal T-admissible sets could exist,
making it unclear which to select to maximize the predictive expressiveness.
However, the next result says that this is not the case, since we can establish the uniqueness of the maximal T-admissible set.

\begin{restatable}[Uniqueness of Maximal T-Admissible Set]{theorem}{uniquenessmaxt}
\label{prop:uniqueness}
For the queries $Q(\rmW_\star)$, a maximal T-admissible set is unique and can be written as:
\begin{equation}
\text{Max-T-Ad}(\rmW_\star) = \cap_{\rmW_i\in \rmW_\star}
\left(\rmW_i\cup ND(\rmW_i)\right).
\end{equation}
Also, $\rmT \in \text{T-Ad}(\rmW_\star)$ if and only if $\rmT\subseteq \text{Max-T-Ad}(\rmW_\star)$ .
\end{restatable}

To illustrate, for the group of queries $Q(\*W_1), Q(\*W_2), \cdots$, the maximal T-admissible set is unique and it is the intersection of non-descendants of $\*W_i$ plus $\*W_i$.
Interestingly, 
identifying a maximal T-admissible set only requires the descendants of $\rmW$ and does not rely on the full specification of the causal graph.
For example, given the features \{cheekbone, smiling, gender\} and the query ``What if the person had smiled?'', it only requires the knowledge of descendants of "smiling", which is ``cheekbone''.
This does not rely on the full latent causal graph, which is often challenging to obtain.

\xadd{An important practical implication of \Cref{thm:graphicalcriteria,prop:uniqueness} is that, given a query $Q(\rmW)$, one could incorporate additional features as long as they are non-descendants of $\rmW$, which would help improve accuracy while retaining the causal interpretability w.r.t. $P(\hY_{\rmW}\mid \rmX)$. For example, given the T-admissible set \{smiling, gender\} and the query ``Would the person be attractive had they smiled?'', one can incorporate additional features, e.g., age or hair color, that are non-descendants of smiling.}

So far, we have described how to find causally interpretable models that can answer counterfactual queries. 
We now describe a practical way of evaluating such queries from the data.
\begin{restatable}[Closed Form]{theorem}{closedform}
\label{thm:closed-form}
If $\Omega_{\textnormal{GCP}(\rmT)}$ is causally interpretable w.r.t. $Q(\rmW)$, the following holds:
\begin{equation}
\label{eq:closed-form}
P(\hY_{\rvw'}\mid \rvx) = \sum_{\rvt} P(\hY \mid \rvw'\cap \rmT, \rvt\setminus \rmW) P(\rvt \mid \rvx).
\end{equation}
\end{restatable}
This implies that the counterfactual quantity can be elicited from a two-step prediction -- (1) a classifier $P(\hY\mid \rmT)$ and (2) a feature extractor $P(\rmT \mid \rmX)$. For example, $Q(S)$ introduced in \Cref{ex:bp-noninterpretability} can be computed using observational data and the maximal T-admissible set \{S, F\} as:
$P(\widehat{Y}_{S=0} \mid \*X) = \sum_{s, f}P(\widehat{Y} \mid S=0, f)P(s, f \mid \*X)$.
Specifically, $\{S, F\}$ are extracted from $P(S, F \mid \rmX)$ and the prediction is made by classifying $P(\widehat{Y} \mid S=0, F)$, conditioning $S=0$.
Note that \Cref{eq:closed-form} only holds when the model is causally interpretable, and it does not hold for non-interpretable ones.

\begin{figure}
\centering
\includegraphics[width=0.95\linewidth]{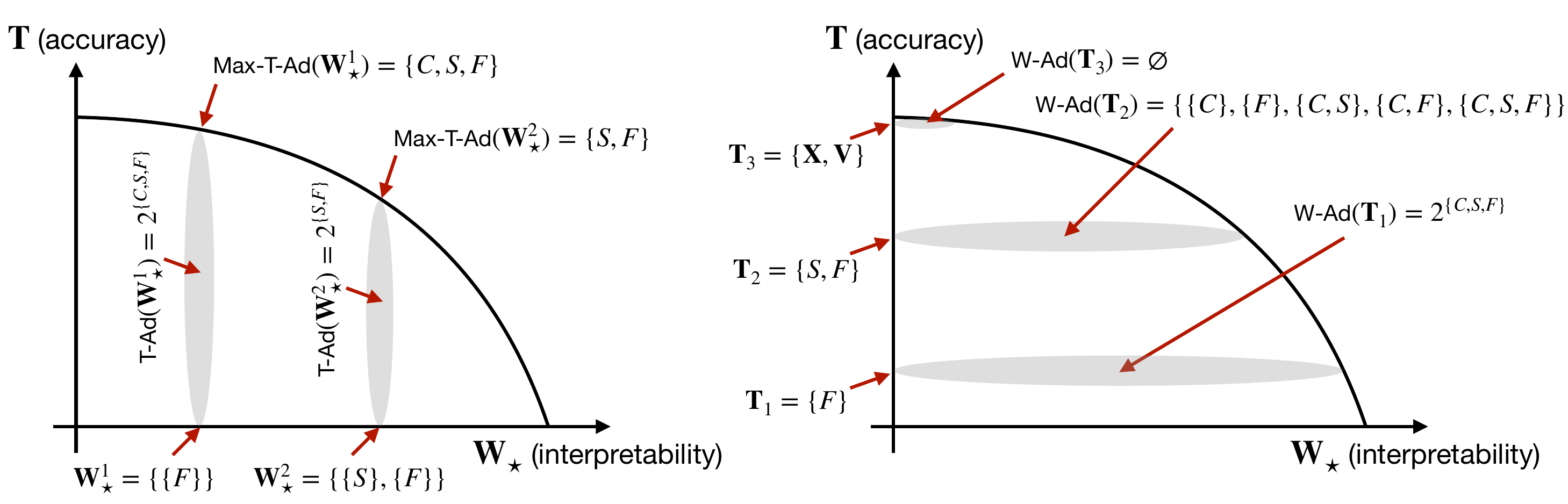}
\caption{(\textbf{Left}) As we want a model to answer more counterfactual queries ($\rmW_\star^1\subseteq \rmW_\star^2$), the predictive power would decrease ($\text{Max-T-Ad}(\rmW_\star^2)\subseteq \text{Max-T-Ad}(\rmW_\star^1)$). (\textbf{Right}) As the predictive power increases ($\rmT_1\subseteq \rmT_2$), interpretable counterfactuals 
would decrease ($\text{W-Ad}(\rmT_2)\subseteq \text{W-Ad}(\rmT_1)$).}
\label{fig:trade-off}
\vspace{-10pt}
\end{figure}

\subsection{Fundamental Trade-Off between Causal Interpretability and Accuracy}

So far, we have developed the machinery for building causally interpretable models that can answer counterfactual queries.
Now, we discuss which queries can be read from the given predictive model architecture. The following formalizes such notions of admissible queries.

\begin{definition}[W-Admissible Set]
\label{def:w-ad}
We say $\rmW$ is W-admissible w.r.t. $\rmT$ if $\Omega_{\textnormal{GCP}(\rmT)}$ is causally interpretable w.r.t. $Q(\rmW)$. A set of W-admissible sets w.r.t. $\rmT$ is denoted as $\text{W-Ad}(\rmT)$.
\end{definition}

For example, in \Cref{fig:comparison-cp},
CP model that uses the features $\{F, S, C\}$ as the predictor of the label can answer counterfactual queries $Q(\{F\}), Q(\{C\}), Q(\{F, S\}), Q(\{F, C\})$ and $Q(\{F, S, C\})$, i.e.,  $\text{W-Ad}(\{F, S, C\}) = \{\{F\}, \{C\}, \{F, S\}, \{F, C\}, \{F, S, C\}\}$ by applying \Cref{thm:graphicalcriteria}.
Similarly, in \Cref{fig:comparison-gcp}, we have $\text{W-Ad}(\{S, C\}) = \{\{F\}, \{S\}, \{C\}, \{F, S\}, \{F, C\}, \{S, C\}, \{F, S, C\}\}$.
Here, one might notice that the model using a larger set of features can answer a smaller number of counterfactual questions. Our next result establishes a trade-off between accuracy and \xadd{causal} interpretability.

\begin{restatable}[\xadd{Causal} Interpretability-Accuracy Trade-Off]{theorem}{tradeoff}
\label{thm:tradeoff}
The following holds: \\
(i) If $\rmT_1\subseteq \rmT_2$, then $\text{W-Ad}(\rmT_2)\subseteq \text{W-Ad}(\rmT_1)$. \\
(ii) If $\rmW_\star^1\subseteq \rmW_\star^2$, then 
$\text{Max-T-Ad}(\rmW_\star^2)\subseteq \text{Max-T-Ad}(\rmW_\star^1)$.
\end{restatable}

In other words, \Cref{thm:tradeoff}-(i) states that the counterfactuals that can be evaluated from the model decrease ($\text{W-Ad}(\rmT_2)\subseteq \text{W-Ad}(\rmT_1)$) as the predictors increase ($\rmT_1\subseteq \rmT_2$). 
Similarly, \Cref{thm:tradeoff}-(ii) states that the predictive power would decrease ($\text{Max-T-Ad}(\rmW_\star^2)\subseteq \text{Max-T-Ad}(\rmW_\star^1)$) as we want the models to answer more counterfactual queries ($\rmW_\star^1\subseteq \rmW_\star^2$).
This reveals a fundamental trade-off between causal interpretability and accuracy, where better predictive power would compromise the interpretability, and vice versa, as illustrated in \Cref{fig:trade-off}.

\section{Experiments}
\label{sec:experiment}

In this section, we evaluate our framework for estimating counterfactuals and compare it with prior approaches. 
Experimental details and additional experimental results are provided in \Cref{appendix:experiment}.

\subsection{Synthetic datasets}
\label{sec:experiment-synthetic}

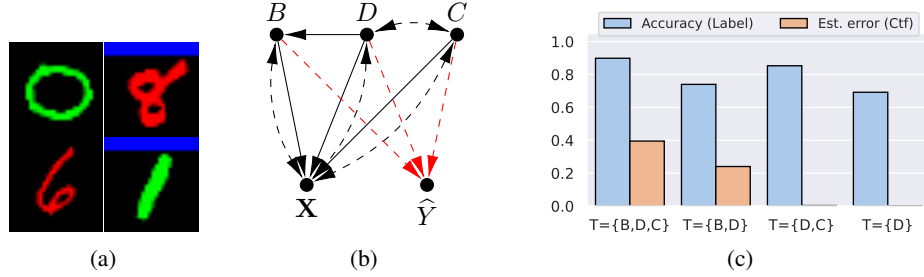
\begin{figure}
\centering
\begin{subfigure}{0.18\linewidth}
\centering
\includegraphics[width=\linewidth]{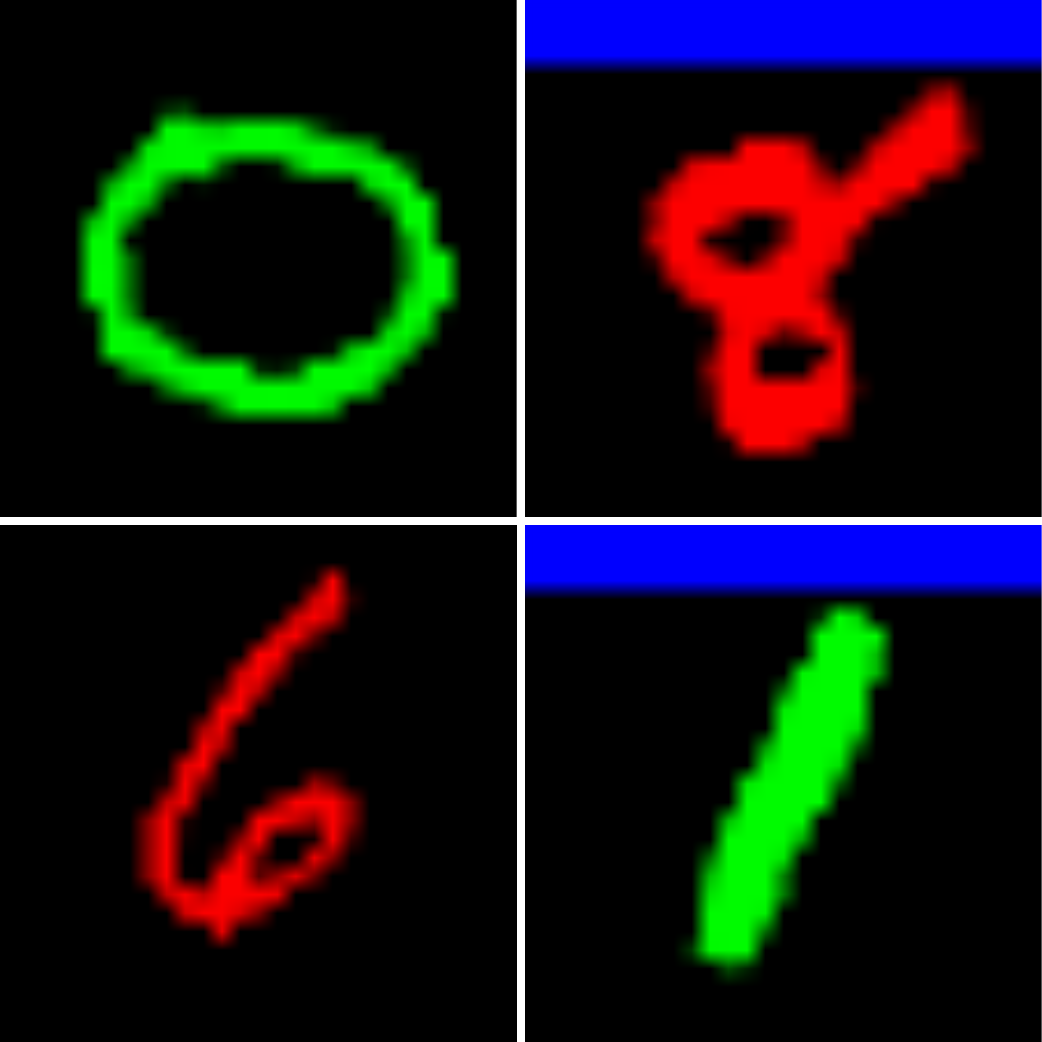}
\caption{}
\label{fig:mnist-dataset}
\end{subfigure}
\hfil
\begin{subfigure}{0.2\linewidth}
\centering
\begin{tikzpicture}[SCM]
\node (1) at (0,1) [label=above:$B$, point];
\node (2) at (1.2,1) [label=above:$D$,point];
\node (3) at (2.4,1) [label=above:$C$, point];
\node (X) at (0.4,-1) [label=below:$\rmX$,point];
\node (Y) at (2,-1) [label=below:$\hY$,point];
\path (2) edge (1);
\path (1) edge (X);
\path (3) edge (X);
\path (2) edge (X);
\path [bd] (2) edge [bend left=20] (3);
\path [bd] (1) edge [bend right=20] (X);
\path [bd] (2) edge [bend left=20] (X);
\path [bd] (3) edge [bend left=20] (X);
\path [redd] (1) edge (Y);
\path [redd] (2) edge (Y);
\path [redd] (3) edge (Y);
\end{tikzpicture}
\caption{}
\label{fig:mnist-cg}
\end{subfigure}
\hfil
\begin{subfigure}{0.4\linewidth}
\centering
\includegraphics[width=0.95\linewidth, trim=0mm 3mm 0mm 0mm, clip]{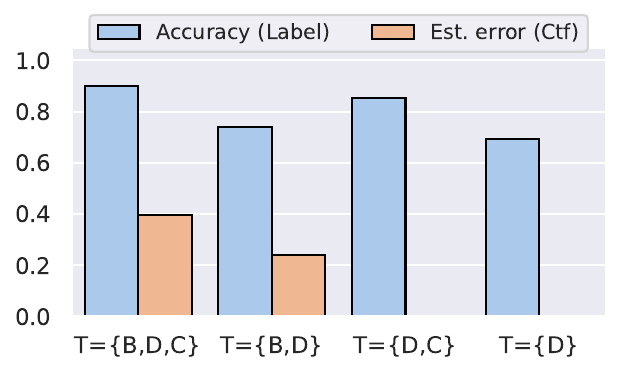}
\caption{}
\label{fig:mnist-tradeoff}
\end{subfigure}
\caption{(a) Example images of BarMNIST dataset. (b) Causal diagram of GCP models. 
Red arrows represent the possible usage for predicting the label.
(c) \xadd{Causal} interpretability-accuracy trade-off.
}
\vspace{-5pt}
\label{fig:mnist-dataset-cg-tradeoff}
\end{figure}

We design the BarMNIST dataset \citep{lecun1998mnist,pan2024counterfactual} where the digits are colored and a bar appears at the top of the image, as shown in \Cref{fig:mnist-dataset}.
Specifically, we consider the features ``bar'' ($B$), ``digit'' ($D$), and ``color'' ($C$), where $D, C$ are correlated and $D$ has a direct causal effect on $B$, as illustrated in \Cref{fig:mnist-cg}. 
The true label is generated from all of the features and unobserved factors.

The dataset allows us to compare the estimation of counterfactuals from each model with the ground-truth.
We trained 4 different models, each using $\rmT = \{B, D, C\}$, $\{B, D\}$, $\{D, C\}$, and $\{D\}$ as the predictor of the label. As shown in \Cref{fig:mnist-tradeoff}, the model using $\rmT = \{B, D, C\}$ achieves the best accuracy, followed by $\rmT = \{B, D\}$ and $\rmT = \{D, C\}$, and the model using $\rmT = \{D\}$ shows the lowest accuracy. 
On the other hand, the best model ($\rmT = \{B, D, C\}$) in terms of accuracy shows a high estimation error on the counterfactual query of changing the digit. \Cref{thm:graphicalcriteria} suggests that any estimation using observed data cannot capture the true counterfactual prediction of this model, since it uses $B$, which is the descendant of $D$. For the same reason, $\rmT = \{B, D\}$ is not causally interpretable, in contrast to $\rmT = \{D, C\}$ and $\rmT = \{D\}$.
Our theory (\Cref{prop:uniqueness}) also suggests that there exists a unique maximal set of features that maintains causal interpretability, in this case, $\rmT = \{D, C\}$.

\begin{figure}[t!]
\centering
\begin{subfigure}{0.48\linewidth}
\centering
\includegraphics[width=\linewidth]{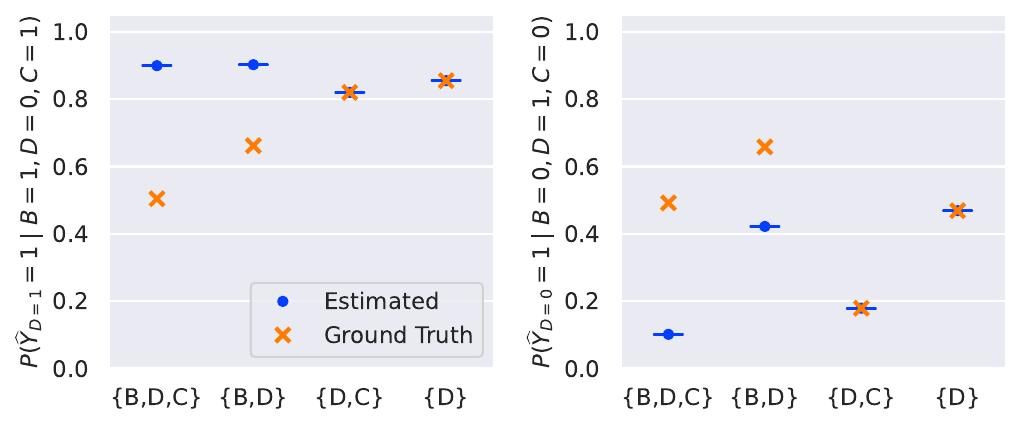}
\caption{Changing digit.}
\label{fig:mnist-ctf-estimation-digit}
\end{subfigure}
\begin{subfigure}{0.48\linewidth}
\centering
\includegraphics[width=\linewidth]{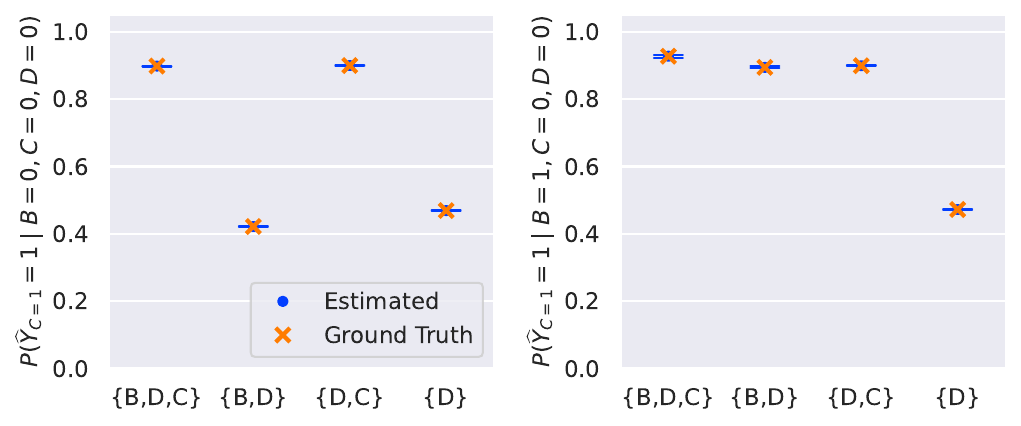}
\caption{Changing color.}
\label{fig:mnist-ctf-estimation-color}
\end{subfigure}
\caption{Estimation of counterfactual queries. Blue dots and orange marks denote estimation of counterfactual queries and ground truth value, respectively.
}
\vspace{-5pt}
\label{fig:mnist-ctf-estimation}
\end{figure}

In \Cref{fig:mnist-ctf-estimation}, we take a closer look at how these models estimate counterfactuals. 
As shown in \Cref{fig:mnist-ctf-estimation-digit}, $\rmT = \{D, C\}$ and $\rmT = \{D\}$ are admissible models for the counterfactual query of changing the digit. On the other hand, for changing color (\Cref{fig:mnist-ctf-estimation-color}), all models are admissible and output a correct estimate of the counterfactual query, since $C$ is not a descendant of any other features.

\begin{figure}[t!]
\centering
\includegraphics[width=0.9\linewidth, trim=0mm 10mm 0mm 0mm, clip]{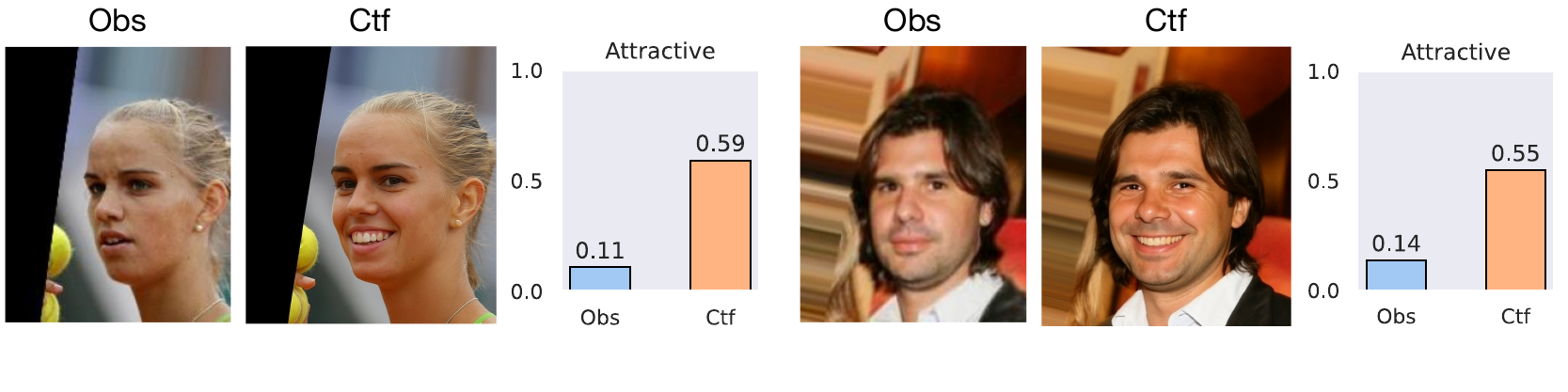}
\caption{Visualization of interpreting counterfactual predictions on CelebA examples.}
\label{fig:celeba-smile}
\vspace{-10pt}
\end{figure}

\subsection{Real-world datasets}
\label{sec:experiment-realworld}

CelebA dataset \citep{liu2018large} contains human face images with the annotations on facial expressions and attributes, such as ``smiling'', ``age'', ``gender'', etc. 
We consider a model predicting the label ``attractiveness'' and examine how a model makes a prediction under counterfactual conditions ``Would the person look attractive had they smiled?''. 
In the real world, it is impossible to observe a counterfactual outcome, but our theory allows us to interpret the behavior of (causally interpretable) models under counterfactual conditions.
Based on \Cref{thm:graphicalcriteria}, we choose the features that are not the descendants of smiling.
\Cref{fig:celeba-smile} illustrates the counterfactual prediction of the model using non-descendant features (i.e., ``smiling'' and ``gender''). We can interpret its behavior under the counterfactual condition that it predicts a higher attractiveness had the one smiled, which is aligned with human common sense.

\section{Conclusion}
\label{sec:conclusion}
In this work, we introduced the notion of causal interpretability, which states whether counterfactual queries can be evaluated from a model and observational data.
By examining commonly used model classes -- blackbox and concept-based models -- we demonstrated that neither is causally interpretable. 
To this end, we developed a graphical criterion that determines whether the model is \xadd{causally} interpretable with respect to the query (\Cref{thm:graphicalcriteria}). We characterize the unique maximal set of features yielding interpretable architecture (\Cref{prop:uniqueness}) and provide a practical way of evaluating such queries from the data (\Cref{thm:closed-form}). Our results reveal a fundamental tradeoff between the causal interpretability and predictive accuracy (\Cref{thm:tradeoff}).
Theoretical findings are corroborated by the experimental results.
\xadd{Additional discussions and limitations are provided in \Cref{appendix:additional-discussions}.}

\begin{ack}
We thank anonymous reviewers for their constructive comments.
This research is supported in part by the NSF, ONR, AFOSR, DoE, Amazon, JP Morgan, and The
Alfred P. Sloan Foundation.
\end{ack}

\bibliographystyle{plainnat}
\bibliography{mainbib}

\newpage
\appendix
\onecolumn
  \DoToC

\section{Proofs and Additional Examples}
\label{appendix:theory}

\subsection{Derivations in Examples}
\label{appendix:theory-example}

\subsubsection{Derivation in Ex.~\ref{ex:bp-noninterpretability}}
In \Cref{ex:bp-noninterpretability}, for the first BP model ${\cM_{\text{BP}}}$, we evaluate $Q(S)$ from $\gM_\text{BP}$ as follows:
\begin{align}
     &P^{\cM_{\text{BP}}}(\widehat{Y}_{S=0} = 1 \mid \mathbf{X}= \mathbf{x}) \nonumber \\
     =& \sum_{f, s, c} P^{\cM_{\text{BP}}}(\widehat{Y}_{S=0} = 1 \mid F=f, S=s, C=c, \mathbf{X}=\mathbf{x})
     P^{\cM_{\text{BP}}}(F=f, S=s, C=c \mid \mathbf{X}=\mathbf{x}) \nonumber\\
    =& P^{\cM_{\text{BP}}}(\widehat{Y}_{S=0}=1 \mid F=0, S=1, C=1, \mathbf{X}=\mathbf{x})
    \nonumber\\
    =& P^{\cM_{\text{BP}}}(f_{\widehat{Y}} \circ f_{\mathbf{X}} (F, S, C, \*U_{\mathbf{X}})_{S=0} = 0 \mid F=0, S=1, C=1, \mathbf{X}=\mathbf{x}) \nonumber \\
    =& P^{\cM_{\text{BP}}}(\mathbf{1}[S > 0.5]_{S=0} = 0 \mid F=0, S=1, C=1, \mathbf{X}=\mathbf{x}), \nonumber 
    \\
    =& \mathbf{1}[S = 0 > 0.5] = 0. \notag
\end{align}
However, for the second BP model, we evaluate $Q(S)$ from $\gM'_\text{BP}$ as:
\begin{align*}
     &P^{\cM_{\text{BP}}'}(\widehat{Y}_{S=0} = 1 \mid \mathbf{X}= \mathbf{x}) \nonumber \\
    =& P^{\cM_{\text{BP}}'}(f'_{\widehat{Y}} \circ f_{\mathbf{X}} (F, S, C, \*U_{\mathbf{X}})_{S=0} = 0 \mid F=0, S=1, C=1, \mathbf{X}=\mathbf{x}) \nonumber \\
    =& P^{\cM_{\text{BP}}'}(\mathbf{1}[U_S > 0.5]_{S=0} = 0 \mid F=0, S=1, C=1, \mathbf{X}=\mathbf{x}) 
    \nonumber \\
     =& \sum_{\*{u}}P^{\cM_{\text{BP}}'}(\mathbf{1}[U_S > 0.5]_{S=0} = 0 \mid \*{u})P^{\cM_{\text{BP}}'}(\*{u} \mid F=0, S=1, C=1, \mathbf{X}=\mathbf{x})) \nonumber & (\text{summing over $\*U$})\\
    =& P^{\cM_{\text{BP}}'}(\mathbf{1}[U_S > 0.5]_{S=0} = 0 \mid U_S = 1) & (S = U_S)\nonumber \\
    =& \mathbf{1}[U_S = 1 > 0.5] = 1. \notag
\end{align*}

\subsubsection{Derivation in Ex.~\ref{ex:cp-noninterpretability}}
In \Cref{ex:cp-noninterpretability}, 
for $\cM_{\text{CP}}$,
\begin{align}
     &P^{\cM_{\text{CP}}}(\widehat{Y}_{S=0} = 1 \mid \mathbf{X}= \mathbf{x}) \nonumber \\
    = &P^{\cM_{\text{CP}}}(\widehat{Y}_{S=0}=1 \mid F=0, S=1, C=1, \mathbf{X}=\mathbf{x}) \nonumber\\
    = &P^{\cM_{\text{CP}}}(\widehat{Y}_{S=0}=1 \mid F=0, S=1, C=1)  &(\widehat{Y} \perp \*X \mid \*V) \nonumber\\
    = &\sum_{c} P^{\cM_{\text{CP}}}(\widehat{Y}_{S=0}=1 \mid C_{S=0}=c)P^{\cM_{\text{CP}}}(C_{S=0}=c \mid F=0, S=1, C=1) \nonumber \\
    = &\sum_{c}P^{\cM_{\text{CP}}}(\widehat{Y}_{S=0}=1 \mid C_{S=0}=c)P^{\cM_{\text{CP}}}(C_{S=0}=c \mid F=0, S=1, C=1) \nonumber \\
    = &P^{\cM_{\text{CP}}}(C_{S=0}=1 \mid F=0, S=1, C=1) &(\text{Eq.~\ref{eq:agg-cp-nonid-1}})\nonumber \\
    =& 0.3 \notag
\end{align}

\subsection{Omitted Proofs}
\label{appendix:theory-proofs}
In this section, we present the proofs of our theoretical results in \Cref{sec:causal-interpretability,sec:generalized}. We first formally introduce the causal diagram induced by an SCM.

\begin{definition}[Causal Diagram {\citep[Def.~13]{bareinboim2022pearl}}]
    \label{def:scm-cg}
    Consider an SCM $\cM = \langle \*U, \*V, \cF, P(\*U)\rangle$. We construct a graph $\cG$ using $\cM$ as follows:
    \begin{enumerate}[label=(\arabic*)]
        \item add a vertex for every variable in $\*V$,
        \item add a directed edge ($V_j \rightarrow V_i$) for every $V_i, V_j \in \*V$ if $V_j$ appears as an argument of $f_{V_i} \in \cF$,
        \item add a bidirected edge ($V_j \leftarrow \rightarrow V_i$) for every $V_i, V_j \in \*V$ if the corresponding $\*U_{V_i}, \*U_{V_j} \subseteq \*U$ are not independent or if $f_{V_i}$ and $f_{V_j}$ share some $U \in \*U$ as an argument.
    \end{enumerate}
    We refer to $\cG$ as the causal diagram induced by $\cM$ (or ``causal diagram of $\cM$'' for short).
    \hfill $\blacksquare$
\end{definition}

We then formally introduce the identifiability of a counterfactual query given an observational distribution and a causal diagram $\cG$.
\begin{definition}[Counterfactual Identification]
\label{def:id}
A counterfactual query $P(y_{1[\*x_1]},y_{2[\*x_2]}, ...)$ is said to be identifiable from $P(\*V)$ and $\cG$, if $P(y_{1[\*x_1]},y_{2[\*x_2]}, ...)$ is uniquely computable from the distributions $P(\*V)$ in any SCM that induces $\cG$.
\end{definition}

Then we start from two lemmas as a tool for the proof of \Cref{thm:graphicalcriteria}. 
\begin{lemma}
\label{lem:inconsistency}
Consider an SCM $\cM$ over $\*V$. Suppose that there exists a path made entirely of bi-directed edges between $V_i, V_j \in \*V$ in $\cG$.
Consider two sets $\*A, \*B \subseteq \*V$ and $\*A \cap \*B = \emptyset$. 
Let the intervened values are not consistent with the factual values, namely, $\*b \not \subseteq \*v$.
Then the query $P(\*a_{\*b} \mid \*v)$ is identifiable from $P(\*V)$ and $\cG$ if and only if $\*A \subseteq ND(\*B)$, where $ND(\*B) = \cap_{B_i\in\*B} ND(B_i)$. 
\end{lemma}
\begin{proof}
($\Rightarrow$) Suppose $\*A \subseteq ND(\*B)$. 
We have 
$P(\*a_{\*b} \mid \*v) = P(\*a \mid \*v) = \mathbf{1}[\*a = \*v]$
which implies that $P(\*a_{\*b} \mid \*v)$ is uniquely computable. 

($\Leftarrow$) Suppose there exists $A \in \*A$ such that $A \in Des(B)$. 
By \citet[Thm.~3]{correa2021nested}, $P(\*a_{\*b} \mid \*v)$ is an inconsistent factor since $\*B \subseteq \*V$ and $\*b \subseteq \*v$, and thus, it is not identifiable from $P(\rmV)$.
\end{proof}

\begin{lemma}[{\citet[Lemma.~1]{correa2021nested}}]
\label{lem:nonid-sum}
Consider an SCM over $\*V$ induce observational distribution $P(\*V)$ and diagram $\cG$. 
Suppose $A_2$ takes input as $A_1$.
Then $\sum _{a_1}P(A_{1[\*b_1]}, A_{2[\*b_2]}, ...)$ is identifiable if and only if $P(A_{1[\*b_1]}, A_{2[\*b_2]}, ...)$ is identifiable. 
\end{lemma}

Now, we are ready to proceed to the proof of \Cref{thm:graphicalcriteria}.

\graphicalcriteria*
\begin{proof}
According to \Cref{def:causal-interpretability,def:generalized-concept-model,def:id}, this is equivalent to prove that query $P(\widehat{y}_{\*w'} \mid \*x)$ is identifiable iff $\*T \subseteq ND(\*W)\cup \*W$ given the observational distribution $P(\*V,\*X, \widehat{Y})$ and the diagram  $\cG^{\mathrm{Aug}}$ over $\{\*V, \*X, \widehat{Y}\}$ (shown in \Cref{fig:proof-diagram}). To illustrate, the diagram $\cG$ over $\*V$ is an arbitrary given DAG; for any $V_i \in \*V$, $V_i$ point to 
$X$ and bi-directed connected to $X$; only a subset $\*T \subseteq \*V$ point to $\widehat{Y}$. Denote $\*Z = \*T\backslash \*W$.
 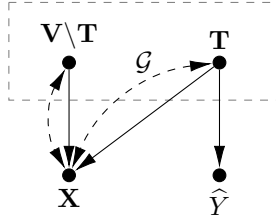
\begin{figure*}[t!]
 \centering
	\begin{subfigure}[b]{0.25\linewidth}
    \centering
    \begin{tikzpicture}[SCM]
    	\filldraw [fill=white, draw=gray,dashed] (-1.8,0.8) rectangle (1.8, -0.5);
        \node (D) at (-1, 0) [label=above:$\*V\backslash\*T$,point];
        \node (C) at (1,0) [label=above:$\*T$, point];
		\node (X) at (-1,-1.5) [label=below:$\*X$,point];
        \node (Y) at (1,-1.5) [label=below:$\widehat{Y}$,point];

        \node (G) at (0,0) {$\cG$};
      
        \path [bd] (D) edge [bend right=30] (X);
        \path [bd] (C) edge [bend right=30] (X);
        \path (D) edge (X);
        \path (C) edge (Y);
        \path (C) edge (X);
		                       
    \end{tikzpicture}
    \label{fig:proof-graph}
  \end{subfigure}
  \vspace{-10pt}
  \caption{Diagrams used in the proof of \Cref{thm:graphicalcriteria}.}
  \label{fig:proof-diagram}
\end{figure*}
\begin{align}
    &P(\widehat{y}_{\*w'} \mid \*x)  \nonumber\\
    =&\sum_{\*v}P(\widehat{y}_{\*w'} \mid \*v, \*x)P(\*v \mid \*x) &(\text{summing over }\*V) \notag \\
    =& \sum_{\*v, \*t''}P(\widehat{y}_{\*w'} \mid \*t''_{\*w'}, \*v, \*x)P(\*t''_{\*w'} \mid \*v, \*x)P(\*v \mid \*x) &(\text{summing over }\*T_{\*w'} \text{ in } \cM_{\*w'} \text{world}) \notag \\
    =& \sum_{\*v, \*t''}P(\widehat{y}_{\*w'} \mid \*t''_{\*w'})P(\*t''_{\*w'} \mid \*v, \*x)P(\*v \mid \*x)  
    &(\hY_{\*w'} \perp \{\*V, \*X\} \mid \rmT_{\*w'}) \label{eq:ind-tm1}\\
    =& \sum_{\*v, \*z''}P(\widehat{y}_{\*w'} \mid \*z''_{\*w'})P(\*z''_{\*w'} \mid \*v, \*x)P(\*v \mid \*x)  &(\text{consistency}) \label{eq:consistency}\\
    =& \sum_{\*v, \*z''}P(\widehat{y} \mid \*z'', \*w')P(\*z''_{\*w'} \mid \*v, \*x)P(\*v \mid \*x)  & (\text{do-calculus \citep{pearl:95a}}) \label{eq:factorization}
\end{align}
\Cref{eq:ind-tm1} holds since $\hY_{\*w}$ are independent with $\*X$ and $\*V$ since all parents of $\hY_{\*w'}$ (which is $\*T_{\*w'}$) are conditioned on.
\Cref{eq:consistency} holds since the $\*T \cap \*W$ should be consistent with the intervened value in $\*w$ (and the remaining variables $\*Z$ in $\*T$ taking $\*z''$. \Cref{eq:factorization} holds due to  $\widehat{Y} \perp \*W \mid \*T$ in $\cG_{\underline{\*W}}$, where $\cG^{}_{\underline{\*W}}$ is the graph removing outgoing edge of $\*W$. Using do-calculus, we have:
\begin{equation}
    P(\widehat{y}_{\*w'} \mid \*z''_{\*w'}) = P(\widehat{y} \mid \*z'', \*w').
\end{equation}
We will prove that \Cref{eq:factorization} is identifiable if and only if $\*T \subseteq ND(\*W)\cup \*W$, which is equivalent to prove \Cref{eq:factorization} is identifiable iff $\*Z \subseteq ND(\*W)$ since $\*Z =  \*T \setminus \*W$. 
According to \Cref{eq:factorization}, the only undermined term is $P(\*z''_{\*w'} \mid \*v, \*x)$. Since $\*V$ and $\*X$ are bi-directly connected, \Cref{lem:inconsistency} suggests $P(\*z''_{\*w'} \mid \*v, \*x)$ is identifiable iff $\*Z \subseteq ND(\*W)$. Then,  $P(\widehat{y} \mid \*z'', \*w')P(\*z''_{\*w'} \mid \*v, \*x)P(\*v \mid \*x)$ is identifiable iff $\*Z \subseteq ND(\*W)$. According to \Cref{lem:nonid-sum}, \Cref{eq:factorization} is identifiable iff $\*T \subseteq ND(\*W)\cup \*W$.
\end{proof}

 \begin{figure*}[t!]
 \centering
	\begin{subfigure}[b]{0.25\linewidth}
    \centering
    \begin{tikzpicture}[SCM]
    	\filldraw [fill=white, draw=gray,dashed] (-1.8,0.8) rectangle (1.8, -0.5);
        \node (D) at (-1, 0) [label=above:$\*V$,point];
		\node (X) at (1,0) [label=above:$\*X$,point];
        \node (Y) at (1,-1.5) [label=below:$\widehat{Y}$,point];

        \node (G) at (0,-0.75) {$\cG$};
      
        \path [bd] (D) edge [bend left=30] (X);
        \path (D) edge (X);
        \path (X) edge (Y);
		                       
    \end{tikzpicture}
    \label{fig:bp-proof-graph}
  \end{subfigure}
  \vspace{-10pt}
  \caption{Diagrams used in the proof of \Cref{prop:bp-noninterpretability}.}
  \label{fig:bp-proof-diagram}
\end{figure*}
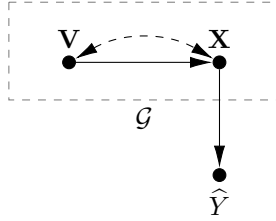
\bpnoninterpretability*
\begin{proof}
Since the observational $P(\mathbf{X})$ is identifiable, we will prove that $P(\widehat{y}_{\*w'}, \*x)$ is not identifiable given a blackbox model structure and observational distribution $P(\mathbf{X}, \widehat{Y}, \mathbf{V})$.
\begin{align}
    &P(\widehat{y}_{\*w'}, \*x)  \nonumber\\
    =&\sum_{\*x'}P(\widehat{y}_{\*w'}, \*x'_{\*w'}, \*x) &(\text{summing over }\*X_{\*w'}) \notag \\
    =& \sum_{\*x'}P(\widehat{y}_{\*x'})P(\*x'_{\*w'}, \*x) &(\text{consistency} \text{ and } \hY_{\*x'} \perp \{\*X, \*X_{\*w'}\}) \notag \\
    =& \sum_{\*x'}P(\widehat{y}_{\*x'})\sum_{\*w} P(\*x'_{\*w'}, \*x, \*w) &(\text{summing over $\*W$}) \notag\\
\end{align}
$ P(\*x'_{\*w'}, \*x, \*w)$ is not identifiable according to Lemma \ref{lem:inconsistency}. Then Lemma \ref{lem:nonid-sum} suggests that $P(\widehat{y}_{\*w'}, \*x)$ is not identifiable.
\end{proof}

\uniquenessmaxt*
\begin{proof}
(i) First, we will show that 
$\rmS \coloneq \cap_{\rmW_i\in \rmW_\star} \left(\rmW_i\cup ND(\rmW_i)\right)$ is a T-admissible set w.r.t $Q(\rmW_\star)$.
For each $\rmW_i\in \rmW_\star$, we have
\begin{equation*}
\cap_{\rmW_i\in \rmW_\star} \left(\rmW_i\cup ND(\rmW_i)\right) \subseteq \rmW_i\cup ND(\rmW_i).
\end{equation*}
Therefore, by \Cref{thm:graphicalcriteria}, $\cap_{\rmW_i\in \rmW_\star} \left(\rmW_i\cup ND(\rmW_i)\right)$ is a T-admissible set w.r.t $Q(\rmW_i)$ for all $\rmW_i\in \rmW_\star$.
Thus, we have $\rmS \in \text{T-Ad}(\rmW_\star)$.

(ii) Now, we will show that $\rmS$ is a maximal T-admissible set w.r.t $\rmW_\star$.
Suppose there exists $\rmS'$ such that $\rmS' \in \text{T-Ad}(\rmW_\star)$ and $\rmS' \supsetneq \rmS$.
Since $\rmS' \in \text{T-Ad}(\rmW_\star)$, $\rmS' \in \text{T-Ad}(\rmW_i)$ for all $\rmW_i\in \rmW_\star$.
Hence, \[\rmS' \subseteq \rmW_i\cup ND(\rmW_i) \quad \text{ for all } \rmW_i\in \rmW_\star.\]
Therefore, $\rmS' \subseteq \cap_{\rmW_i\in \rmW_\star} \left(\rmW_i\cup ND(\rmW_i)\right) = \rmS$, which contradicts $\rmS' \supsetneq \rmS$. Therefore, $\rmS$ is a maximal T-admissible set w.r.t $\rmW_\star$.

(iii) Now, we will show that $\rmS$ is a unique maximal T-admissible set.
Suppose there exists another maximal T-admissible set $\rmS'$. Since $\rmS' \in \text{T-Ad}(\rmW_\star)$, we have $\rmS' \subseteq \rmS$ by the same reason in (ii). If $\rmS' \subsetneq \rmS$, then it contradicts that $\rmS'$ is a maximal T-admissible set, since $\rmS$ is a T-admissible set. Therefore, we have $\rmS = \rmS'$.
In other words, a maximal T-admissible set is unique and can be written as $\text{Max-T-Ad}(\rmW_\star) = \cap_{\rmW_i\in \rmW_\star} \left(\rmW_i\cup ND(\rmW_i)\right)$.

(iv) Now, we will show that $\rmT \in \text{T-Ad}(\rmW_\star)$ if and only if $\rmT\subseteq \text{Max-T-Ad}(\rmW_\star)$.
Suppose $\rmT \in \text{T-Ad}(\rmW_\star)$. Then, by (ii), we have $\rmT\subseteq \cap_{\rmW_i\in \rmW_\star} \left(\rmW_i\cup ND(\rmW_i)\right)$. Also, we showed that $\text{Max-T-Ad}(\rmW_\star) = \cap_{\rmW_i\in \rmW_\star} \left(\rmW_i\cup ND(\rmW_i)\right)$. Therefore, we have $\rmT\subseteq \text{Max-T-Ad}(\rmW_\star)$. Now, suppose that $\rmT\subseteq \text{Max-T-Ad}(\rmW_\star)$. We have $\rmT \subseteq \cap_{\rmW_i\in \rmW_\star} \left(\rmW_i\cup ND(\rmW_i)\right)$, and thus, $\rmT \subseteq \rmW_i\cup ND(\rmW_i)$ for all $\rmW_i\in \rmW_\star$. Therefore, $\rmT \in \text{T-Ad}(\rmW_i)$ for all $\rmW_i\in \rmW_\star$, and thus, $\rmT \in \text{T-Ad}(\rmW_\star)$.
\end{proof}

\closedform*
\begin{proof}
From \Cref{eq:factorization}, we have
\begin{equation}
    P(\widehat{y}_{\*w} \mid \*x)  = \sum_{\*v, \*z''}P(\widehat{y} \mid \*z'', \*w')P(\*z''_{\*w'} \mid \*v, \*x)P(\*v \mid \*x).
\end{equation}
Note that this equation is identifiable if only if $\*Z \subseteq  \*W \cup {ND}(\*W)$. Then
\begin{align}
    &= \sum_{\*v, \*z''}P(\widehat{y} \mid \*z'', \*w')P(\*z''_{\*w'} \mid \*v, \*x)P(\*v \mid \*x) \nonumber\\
    &= \sum_{\*v, \*z''}P(\widehat{y} \mid \*z'', \*w')\mathbf{1}[\*z'' = \*v]P(\*v \mid \*x) & \text{(Lemma.~\ref{lem:inconsistency})} \nonumber\\
    &= \sum_{\*v}P(\widehat{y} \mid \*t\setminus\*w, \*w')P(\*v \mid \*x)   &\text{(where $\*z = (\*t\setminus\*w) \in \*v$)} \nonumber\\
    &= \sum_{\*v}P(\widehat{y} \mid \*t\setminus\*w, \*w'\cap \*t)P(\*v \mid \*x)  & (\*Y \perp \*W \setminus \*T \mid \*T) \nonumber\\
        &= \sum_{\*t}P(\widehat{y} \mid \*t\setminus\*w, \*w'\cap \*t)P(\*t \mid \*x). \label{eq:final}
\end{align}
This conclude $P(\widehat{Y}_{\*w} \mid \*x) = \sum_{\*t}P(\widehat{Y} \mid \*w'\cap \*T, \*t\setminus\*W)P(\*t \mid \*x)$ since Eq.~\ref{eq:final} holds for any $\rvt,\rvw$.
\end{proof}

\tradeoff*
\begin{proof}
(i) Let $\rmT_1 \subseteq \rmT_2$.
Suppose $\rmW \in \text{W-Ad}(\rmT_2)$. By \Cref{def:w-ad} and \Cref{thm:graphicalcriteria}, we have 
\[
    \rmT_2 \subseteq \rmW\cup ND(\rmW).
\]
Since $\rmT_1 \subseteq \rmT_2$, it follows that $\rmT_1 \subseteq \rmW\cup ND(\rmW)$. Therefore, by \Cref{def:w-ad} and \Cref{thm:graphicalcriteria}, $\rmW \in \text{W-Ad}(\rmT_1)$. 
Thus, for all $\rmW \in \text{W-Ad}(\rmT_2)$, we have $\rmW \in \text{W-Ad}(\rmT_1)$.
Hence, we have
\[
    \text{W-Ad}(\rmT_2) \subseteq \text{W-Ad}(\rmT_1).
\]

(ii) Let $\rmW_\star^1\subseteq \rmW_\star^2$.
Then, we have 
\[
    \cap_{\rmW_i\in \rmW_\star^2} \left(\rmW_i\cup ND(\rmW_i)\right) \subseteq \cap_{\rmW_i\in \rmW_\star^1} \left(\rmW_i\cup ND(\rmW_i)\right).
\]
Therefore, we have $\text{Max-T-Ad}(\rmW_\star^2) \subseteq \text{Max-T-Ad}(\rmW_\star^1)$ by \Cref{prop:uniqueness}.
\end{proof}

\subsection{Additional Examples}

The following example illustrates how GCP and CP models compare.

\begin{example}[GCP]
\label{ex:gcp}
Consider the generative process of observed concepts $\*V_0 = \{F, S, C\}$ and the image $\*X$, as in \Cref{ex:bp} (BP model) and \Cref{ex:cp} (CP model). 
Consider a GCP model $\cM_{\text{GCP}} = \langle \*U=\{U_F, U_{S}, U_{C_1}, U_{C_2}, \*U_{\*X}\}, \{\{F, S, C\}, \*X, \widehat{Y}\}, 
 \cF^{\text{GCP}}, P^{\text{GCP}}(\*U) \rangle$, where
\begin{equation}
\label{eq:gcp}
{\cF^{\text{GCP}}} = \left\{
\begin{aligned}
F & \leftarrow U_F \oplus U_S \\
S & \leftarrow U_S \\
C & \leftarrow (\neg S \land U_{C_1}) \oplus (S \land U_{C_2})  \\
\*X & \leftarrow {f}_{\*X}(F, S, C, \*U_{\*X}) \\
{\widehat{Y}} & \leftarrow \textcolor{red}{f^{\text{GCP}}_{\widehat{Y}}(S, F)}
\end{aligned}
\right.
\end{equation}
and $P^\text{GCP}(\*U)$ is equal to $P^\text{CP}(\*U)$ in \Cref{ex:cp}. The causal diagram induced by GCP model $\cM_{\text{GCP}}$ is shown in \Cref{fig:comparison-gcp}.
To illustrate, instead of predicting the label based on pixels in images $\*X$ (BP models) or 
all observed features $\{F, S, C\}$ (CP models), GCP model makes a prediction using a selected subset of features $\*T = \{S, F\}$ (i.e., smiling and gender) in this case. \hfill $\blacksquare$
\end{example}

\begin{wraptable}{r}{0.4\textwidth}
\centering
\vspace{-10pt}
\begin{adjustbox}{width=0.75\linewidth}%
\begin{tabular}{cccc}
\toprule
$F$ & $S$ & $C$ & $P(F, S, C)=1$  \\ 
\midrule
0     & 0     & 0    & 0.168             \\ %
0     & 0     & 1    & 0.072       \\ %
0     & 1     & 0    & 0.096           \\ %
0     & 1     & 1    & 0.144				\\ %
1     & 0     & 0    & 0.112           \\ %
1     & 0     & 1    & 0.048            \\ %
1     & 1     & 0    & 0.144          \\ %
1     & 1     & 1    & 0.216 \\
\bottomrule
\end{tabular}
\end{adjustbox}
\caption{Probability table in \Cref{ex:gcp-id}.}
\label{tab:gcp-id-obs}
\vspace{-10pt}
\end{wraptable}

The following example illustrates the case where the GCP model is causal interpretable.
\begin{example}[Continued from \Cref{ex:gcp}]
\label{ex:gcp-id}
Consider $\Omega_{\textnormal{CP}}$ in \Cref{ex:cp-noninterpretability}. \Cref{thm:graphicalcriteria} suggests $\Omega_{\textnormal{CP}}$ is not interpretable w.r.t. to query $Q(S)$ $P(Y_{S=0} \mid \*X)$.
This is because 
$C\in \text{De}(S)$, where $\rmW=\{S\}$, i.e., the prediction of $\widehat{Y}$ is made based on $C$, a descendant of $S$. 
In contrast, $\Omega_{\textnormal{GCP}(\{S, F\})}$ in \Cref{ex:gcp} is said to be causally interpretable w.r.t. to query $P(Y_{S=0} \mid \*X)$ since ${f}^{\text{GCP}}_\hY$ only takes $\rmT=\{S, F\} \subseteq S \cup ND(S)$ as input. 
To illustrate, let us consider the GCP model $\cM_\text{GCP}$ in \Cref{ex:gcp}. Similar to \Cref{ex:bp-noninterpretability,ex:cp-noninterpretability}, let the observational quantity $P(F=0, S=1, C=1 \mid \mathbf{X}={\mathbf{x}}) = 1$ and let $f_{\widehat{Y}}$ be:
\begin{align}
    \widehat{Y} \leftarrow {f}^{\text{GCP}}_{\widehat{Y}} (S, F) = \mathbf{1}[S + F > 0.5].
\end{align}
Now, consider another GCP model 
\begin{equation}
    \cM_{\text{GCP}}' = \langle \*U'=\{U'_F, U'_{S_1}, U'_{S_1}, U'_{C_1}, U'_{C_2}, \*U'_{\*X}\}, \{\{F, S, C\}, \*X, \widehat{Y}\}, 
 \cF^{\text{GCP}'}, P^{\text{GCP}'}(\*U) \rangle,
\end{equation}
 where
\begin{equation}
{\cF^{\text{GCP}'}} = \left\{
\begin{aligned}
F & \leftarrow U'_F \\
S & \leftarrow ((\neg U'_F) \land U'_{S_1}) \oplus (U'_F \land U'_{S_2}) \\
C & \leftarrow (\neg S \land U'_{C_1}) \oplus (S \land U'_{C_2})  \\
\*X & \leftarrow {f}_{\*X}(F, S, C, \*U_{\*X}) \\
{\widehat{Y}} & \leftarrow \mathbf{1}[S + F > 0.5]
\end{aligned}
\right.
\end{equation}
and $P(U'_F = 1) = 0.52, P(U'_{S_1} = 1) = 0.5, P(U'_{S_2} = 1) = 9/13, P(U'_{C_1} = 1) = 0.5, P(U'_{C_2} = 1) = 0.6$. 
It is verifiable that $P^{\cM_\text{GCP}}(\*V) = P^{\cM_\text{GCP}'}(\*V)$ as shown in \Cref{tab:gcp-id-obs}. Since  $f_{\widehat{Y}}$ is the same in both $\cM_\text{GCP}$ and $\cM'_\text{GCP}$, $P^{\cM_\text{GCP}}(\*V, \widehat{Y}) = P^{\cM_\text{GCP}'}(\*V, \widehat{Y})$.
Let the distribution of $\*U_{\*X}$ satisifies that $P^{\cM_\text{GCP}}(\*V, \*X, \widehat{Y}) = P^{\cM_\text{GCP}'}(\*V, \*X, \widehat{Y})$.
$\cM_{CP}'$ is compatible the graphical constraints induced by the model in \Cref{fig:comparison-cp}. Notice that $f'_{F}, f'_{S}, f'_{C}$ in $\cM_{\text{GCP}}'$ are totally different to $f_{F}, f_{S}, f_{C}$ in $\cM_{\text{GCP}}$.
For the first GCP model $\cM_{\text{GCP}}$,
\begin{align}
     P^{\cM_{\text{GCP}}}(\widehat{Y}_{S=0} = 1 \mid \mathbf{X}= \mathbf{x}) \nonumber 
    = P^{\cM_{\text{GCP}}}(F_{S=0}=1 \mid F=0, S=1, C=1) \nonumber
    = 0.
\end{align}
Similarly, for the second GCP model $\cM_{\text{GCP}}'$,
\begin{align}
     &P^{\cM_{\text{GCP}}'}(\widehat{Y}_{S=0} = 1 \mid \mathbf{X}= \mathbf{x}) \nonumber 
    = P^{\cM_{\text{GCP}}'}(C_{S=0}=1 \mid F=0, S=1, C=1)= 0. \nonumber
\end{align}
This shows that the two GCP models are consistent with the query. 
In other words, if one uses the features $\{S, F\}$ to predict $\widehat{Y}$ , the model architecture in \Cref{fig:comparison-gcp} is guaranteed to provide a unique answer for the counterfactual question "What would the attractiveness prediction be had the person not smiled?" (i.e., $P(Y_{S=0} \mid \*X)$). 
Then one can trust the counterfactual quantities induced by any model with this architecture.
    \hfill $\blacksquare$
\end{example}

\section{Experiments}
\label{appendix:experiment}

In this section, we describe the details for the experiments and provide additional experimental results.

\subsection{Dataset}
\label{appendix:experiment-setup}

\subsubsection{BarMNIST}
\label{appendix:experiment-setup-synthetic1}
For BarMNIST experiment discussed in \Cref{sec:experiment-synthetic}, the data generating process is as follows:
\begin{equation}
{\cF} = \left\{
\begin{aligned}
D & \leftarrow U_D \\
C & \leftarrow U_D \oplus U_C \\
B & \leftarrow (U_{B_1} \wedge D) \oplus (U_{B_1} \land U_{B_2}) \oplus ((\neg U_{B_1}) \land U_{B_2}) \\
\*X & \leftarrow {f}_{\*X}(B, D, C, \*U_{\*X}) \\
Y & \leftarrow \left((D\oplus C) \vee B \right) \oplus U_Y,
\end{aligned}
\right.
\end{equation}
the exogenous variables $U_D, U_C, U_{B_1}, U_{B_2}, U_{B_3}, U_Y$ are independent binary variables, and $P(U_{D} = 1) = 0.5, P(U_{C} = 1) = 0.4, P(U_{B_1} = 1) = 0.9, P(U_{B_2} = 1) = 1/18, P(U_{B_3} = 1) = 0.5, P(U_Y=1) = 0.1$.

Following this process, we generated 60,000 images and corresponding labels, where each image is annotated with 3 binary features, i.e., bar ($B$), color ($C$), and digit ($D$). Here, $D=0$ represents the digits from 0 to 4 and $D=1$ represents the digits from 5 to 9.

\subsubsection{CelebA}
\label{appendix:experiment-setup-celeba}

CelebA dataset \citep{liu2018large} contains 202,599 celebrity facial images, where each image is annotated with 40 different attributes. 
In our experiments, we used the attribute ``attractiveness'' as the label, where the label and all other features are binary.

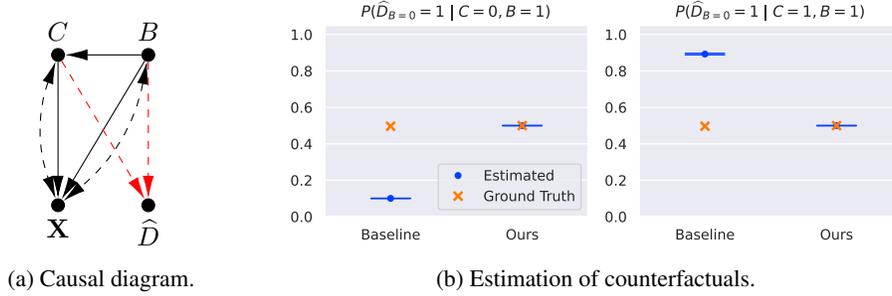
\begin{figure}[t!]
\centering
\begin{subfigure}{0.2\linewidth}
\centering
\begin{tikzpicture}[SCM]
\node (1) at (0,1) [label=above:$C$, point];
\node (2) at (1.2,1) [label=above:$B$,point];
\node (X) at (0,-1) [label=below:$\rmX$,point];
\node (Y) at (1.2,-1) [label=below:$\hD$,point];
\path (2) edge (1);
\path (1) edge (X);
\path (2) edge (X);
\path [bd] (1) edge [bend right=20] (X);
\path [bd] (2) edge [bend left=20] (X);
\path [redd] (1) edge (Y);
\path [redd] (2) edge (Y);
\end{tikzpicture}
\caption{Causal diagram.}
\label{fig:mnist2-cg}
\end{subfigure}
\hfil
\begin{subfigure}{0.6\linewidth}
\centering
\includegraphics[width=\linewidth]{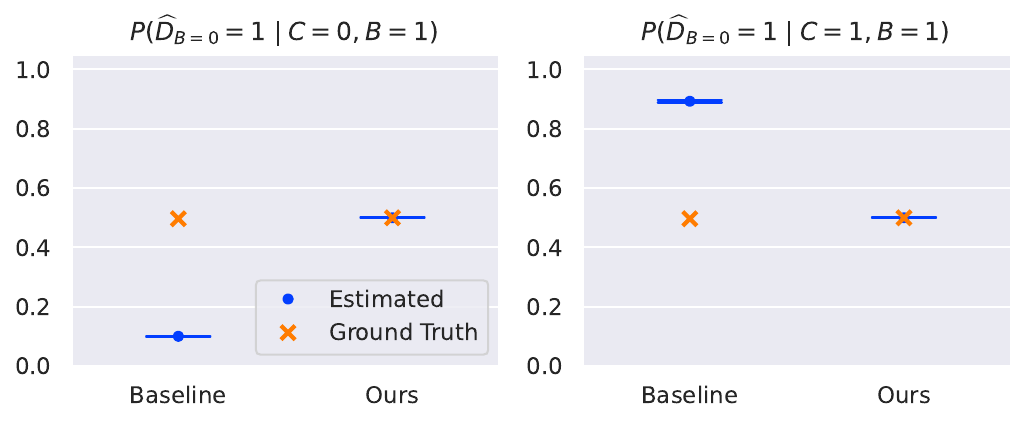}
\caption{Estimation of counterfactuals.}
\label{fig:mnist2-ctf}
\end{subfigure}
\caption{(a) Causal diagram of GCP models. 
Red arrows represent the possible usage for predicting the label.
(b) Estimation of counterfactual queries. Blue dots and orange marks denote estimation of counterfactual queries and ground truth value, respectively.
}
\vspace{-5pt}
\label{fig:mnist2-cg-ctf-tradeoff}
\end{figure}

\subsection{Experimental Details}
In BarMNIST, we used ResNet18 for the feature extractor. For the classifier, we used a three-layer MLP with the hidden dimension of 32 and leakyrelu activation. 
We set the batch size to 1024 and trained the models for 100 epoch. We used Adam optimizer with a learning rate of 0.0003.

In CelebA, we used ResNet34 for the feature extractor and used linear classifier. 
We set the batch size to 512 and trained the models for 100 epochs. We used SGD optimizer with the learning rate of 0.001. We resized the image with center crop into 64$\times$64 for training.

\begin{figure}[t!]
\centering
\includegraphics[width=\linewidth]{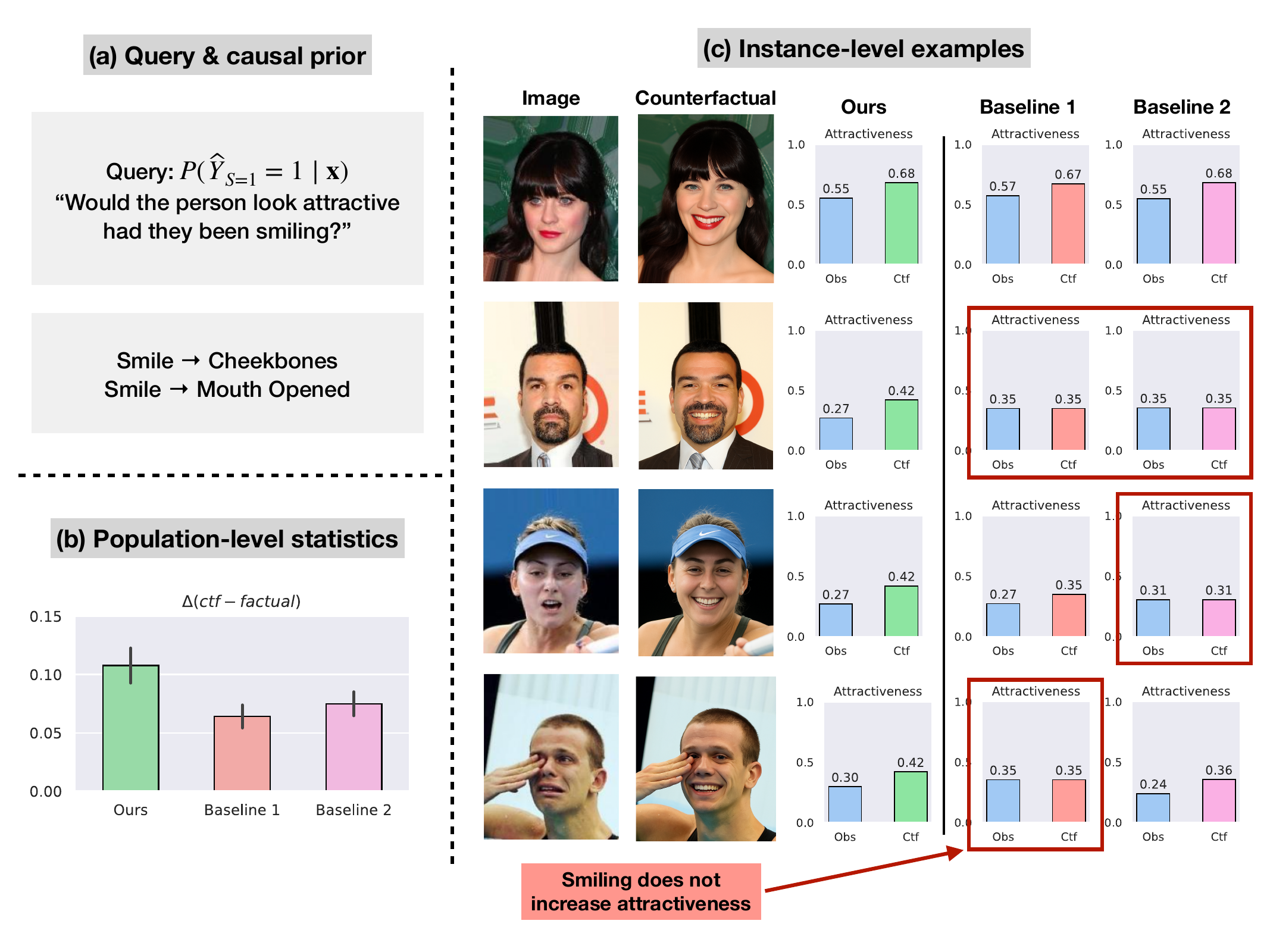}
\caption{(a) We examine the prediction of the models under counterfactual condition. We use causal prior knowledge that smiling has causal effects on the features ``cheekbones'' and ``opened mouth''. 
(b) Average difference between the estimated counterfactual prediction and the prediction on the observed (factual) image.
(c) Qualitative examples for our model and baselines.
}
\label{fig:celeba-detailed-analysis}
\end{figure}

For the training of our model and baselines, we used binary classification loss for both the feature extractor and the classifier, where they are trained simultaneously in an end-to-end manner. 
All experimental results are averaged over 5 independent runs. We report a standard error as the error bar in \Cref{fig:mnist-ctf-estimation,fig:mnist2-cg-ctf-tradeoff,fig:celeba-detailed-analysis}.
All experiments are conducted on a single NVIDIA A100 GPU. 
For the implementation, we utilized publicly available code from \citet{espinosa2022concept}.
We used GPT-4o to generate the counterfactual images shown in \Cref{fig:celeba-smile,fig:celeba-detailed-analysis} to provide an intuitive understanding of the counterfactual questions. 

\subsection{Additional Experimental Results}
\label{appendix:experiment-additional}

\subsubsection{BarMNIST}

To validate our theory with a different graph structure, we consider a causal diagram in \Cref{fig:mnist2-cg} where the goal is to predict the digit $D$ from the image. 
The data generating process is as follows:
\begin{equation}
{\cF} = \left\{
\begin{aligned}
B & \leftarrow U_B \\
C & \leftarrow B \vee U_{C_1} \oplus U_{C_2}\\
D & \leftarrow (B \vee C) \oplus U_D \\
\*X & \leftarrow {f}_{\*X}(B, D, C, \*U_{\*X}),
\end{aligned}
\right.
\end{equation}
where the exogenous variables $U_B, U_{C_1}, U_{C_2}, U_D$ are independent binary variables, where $P(U_B=1)=0.6, P(U_{C_1}=1)=0.5, P(U_{C_2}=1)=0.1, P(U_D=1)=0.1$.

The baseline model uses the features $B$ and $C$ for predicting the label, and our model uses $B$ for making a prediction. Our theory (\Cref{thm:graphicalcriteria}) suggests that our model is causally interpretable, but not the baseline which uses $C$, a descendant of $B$.
We compare our model and baselines for estimating the counterfactual prediction of the model, where the query is to change the bar, i.e., $P(\hD_{B=0} \mid \rvx)$.

\Cref{fig:mnist2-ctf} illustrates the estimation of counterfactual queries (blue dots) and ground truth values (orange marks). This shows that our model correctly estimates counterfactual queries. In contrast, the estimation of the baseline significantly differs from the ground truth. This corroborates our theory that our estimation can properly interpret the counterfactual behavior of the causally interpretable models, but it is not possible for non-interpretable ones.

\subsubsection{CelebA}
Here, we provide a detailed analysis of CelebA experiments in \Cref{sec:experiment-realworld}.
\Cref{fig:celeba-detailed-analysis}-(a) illustrates the counterfactual question and causal prior we utilized to construct our model. Specifically, we leverage the common-sense knowledge that smiling has direct causal influence to the features ``cheekbones'' and ``opened mouth''.
To construct our model, we choose features that are non-descendants of smiling, specifically ``smile'' and ``gender'' as feature set $\rmV$. 
Baselines include descendant features. In \Cref{fig:celeba-detailed-analysis}, baseline 1 uses the features ``smiling'', ``gender'', and ``cheekbones'' and baseline 2 uses the features ``smiling'', ``gender'', ``cheekbones'', and ``opened mouth''.

\Cref{fig:celeba-detailed-analysis}-(b) shows the average difference between the estimated counterfactual prediction and the prediction on the observed image. 
\Cref{fig:celeba-detailed-analysis}-(c) shows qualitative examples comparing our method and baselines. The first column in \Cref{fig:celeba-detailed-analysis}-(c) shows the input image, and the second column illustrates the counterfactual image, as a reference to provide a better understanding of the counterfactual query. 

The theory suggests that a causally interpretable model can properly estimate its prediction under counterfactual conditions. As shown in \Cref{fig:celeba-detailed-analysis}-(b) and (c), our model, which is causally interpretable, consistently increases the attractiveness across the instances, which is also aligned with human reasoning.
In contrast, as illustrated in \Cref{fig:celeba-detailed-analysis}-(c), the estimation of the baselines (which use similar feature set as ours) shows that smiling often does not increase attractiveness (red boxes). In fact, our theory suggests that it is not possible to interpret the counterfactual behavior of non-interpretable models using only observational data, and any attempts to estimate it would lead to inconsistent results.

\section{Additional Discussions, Limitations, and Future Work}
\label{appendix:additional-discussions}

\paragraph{Estimation of the concepts.}
\xadd{
In the closed-form formula in \Cref{eq:closed-form}, the concepts $\mathbf{W}$ and $\mathbf{T}$ are ground-truth concepts. Since the labels of ground-truth concepts are available, one can estimate $P(\mathbf{T} \mid \mathbf{X})$ over ground truth concept $\mathbf{T}$. For clarity, let us denote this estimated distribution as $\widehat{P}(\mathbf{T} \mid \mathbf{X})$.
In the prediction stage, the true concepts $\mathbf{W}$ and $\mathbf{T}$ of an image instance $\mathbf{X}$ are not given directly. Instead, the predicted concepts $\mathbf{W}$ and $\mathbf{T}$ are sampled through the estimated $\widehat{P}(\mathbf{T} \mid \mathbf{X})$. When $\widehat{P}(\mathbf{T} \mid \mathbf{X})$ is accurate, the sampled (predicted) concepts are expected to align closely with the ground-truth concepts. However, if the estimation has an error, the predicted concepts may deviate from the true ones, and this error will naturally propagate into the counterfactual evaluation via \Cref{eq:closed-form}.
}

\xadd{
Our goal with this formulation is to formally characterize how these counterfactual quantities can be computed from the observational distribution under ideal conditions (i.e., accurate estimation). The challenge of robustly estimating $P(\mathbf{T} \mid \mathbf{X})$ from finite data is indeed fundamental and highly relevant to practice, but falls outside the scope of this work. Nevertheless, it would be a valuable direction for future investigation, particularly in light of ongoing research in counterfactual estimation within the causal inference literature \citep{jung2021estimating,jung2023estimating} and the importance of creating more interpretable methods in practice.}

\paragraph{Causal graph.}
Our work reveals that understanding and harnessing causal relationships among the generative features are crucial for building interpretable models that can properly evaluate counterfactual questions. It is important to note that our framework only requires the causal prior on the descendants, and this is a much relaxed assumption compared to the conventional assumption in causal inference, where the full specification of the causal graph is needed \citep{hwang2024positivity,li2024disentangled}.

\paragraph{Real-world datasets.}
\xadd{In real-world datasets, it is infeasible to evaluate the actual value of the counterfactual query because the underlying ground-truth data-generating process for real-world datasets is not given, specifically, the mechanisms of $\mathbf{V}$ are not known. For example, it is unknown how nature decides the generation process of human facial features. Due to this inevitable restriction, we thoroughly validated our theory in BarMNIST datasets (where we have the ground-truth SCM), including causal interpretability-accuracy tradeoff.}

\xadd{Still, our theory allows us to understand the interplay between causal interpretability and accuracy in real-world datasets. For example, given T-admissible set {smiling, gender} and the query ``Would the person be attractive had they smiled?’’, if one wants to incorporate additional query ``Would the person be attractive had they be a men?’’, we know the model using this $\mathbf{T}$ maintains causal interpretability w.r.t. both queries, and thus, would not compromise accuracy.}

\end{document}